\documentclass[preprint,5p,12pt,twocolumn]{article}

\usepackage{amssymb}
\usepackage{subfig}
\usepackage{algorithm,algorithmic}
\usepackage{graphicx}
\usepackage{caption}
\usepackage{xspace}
\usepackage{multirow}
\usepackage{rotating}
 \usepackage{url}
\usepackage{gensymb}
\usepackage [ table ]{ xcolor }
\usepackage[top=2cm, bottom=2cm, left=1.5cm, right=1.5cm]{geometry}
\usepackage{setspace}
\usepackage{morefloats}

\usepackage[affil-it]{authblk}

\begin{document}
\title{Humans and deep networks largely agree on which kinds of variation make object recognition harder}

\author{Saeed Reza Kheradpisheh$ ^{1}$ }
\author{Masoud Ghodrati$ ^{2} $}

\author{Mohammad Ganjtabesh$ ^{1}$}
\author{Timoth\'ee Masquelier$ ^{3,4,5,6,}$\footnote{Corresponding author.\\ Email addresses:\\ kheradpisheh@ut.ac.ir (SRK),\\masoud.ghodrati@monash.edu (MGh) \\ mgtabesh@ut.ac.ir (MG),\\ timothee.masquelier@alum.mit.edu (TM).} }
\affil{\footnotesize $ ^{1} $ Department of Computer Science, School of Mathematics, Statistics, and Computer Science, University of Tehran, Tehran, Iran}
\affil{\footnotesize $ ^{2} $ Department of Physiology, Monash University, Melbourne, VIC, Australia}
\affil{\footnotesize $ ^{3} $ INSERM, U968, Paris, F-75012, France}
\affil{\footnotesize $ ^{4} $ Sorbonne Universit\'es, UPMC Univ Paris 06, UMR-S 968, Institut de la Vision, Paris, F-75012, France}
\affil{\footnotesize $ ^{5} $ CNRS, UMR-7210, Paris, F-75012, France}
\affil{\footnotesize $^{6}$ Centre de Recherche Cerveau et Cognition (CerCo), Centre National de la Recherche Scientifique (CNRS), Universit\'e Paul Sabatier, 31052 Toulouse, France}
\date{}

\maketitle
\begin{abstract}


View-invariant object recognition is a challenging problem, which has attracted much attention among the psychology, neuroscience, and computer vision communities. Humans are notoriously good at it, even if some variations are presumably more difficult to handle than others (e.g. 3D rotations). Humans are thought to solve the problem through hierarchical processing along the ventral stream, which progressively extracts more and more invariant visual features. This feed-forward architecture has inspired a new generation of bio-inspired computer vision systems called deep convolutional neural networks (DCNN), which are currently the best algorithms for object recognition in natural images. Here, for the first time, we systematically compared human feed-forward vision and DCNNs at view-invariant object recognition using the same images and controlling for both the kinds of transformation (position, scale, rotation in plane, and rotation in depth) as well as their magnitude, which we call ``variation level''. We used four object categories: cars, ships, motorcycles, and animals. All 2D images were rendered from 3D computer models. In total, 89 human subjects participated in 10 experiments in which they had to discriminate between two or four categories after rapid presentation with backward masking. We also tested two recent DCNNs (proposed respectively by Hinton's group and Zisserman's group) on the same tasks. We found that humans and DCNNs largely agreed on the relative difficulties of each kind of variation: rotation in depth is by far the hardest transformation to handle, followed by scale, then rotation in plane, and finally position (much easier). This suggests that humans recognize objects mainly through 2D template matching, rather than by constructing 3D object models, and that DCNNs are not too unreasonable models of human feed-forward vision. In addition, our results show that the variation levels in rotation in depth and scale strongly modulate both humans' and DCNNs' recognition performances. We thus argue that these variations should be controlled in the image datasets used in vision research.

\end{abstract}
\section{Introduction}

As our viewpoint relative to an object changes, the retinal representation of the object tremendously varies across different dimensions. Yet our perception of objects is largely stable. How humans and monkeys can achieve this remarkable performance has been a major focus of research in visual neuroscience~\cite{dicarlo2012does}. Neural recordings showed that some variations are treated by early visual cortices, e.g., through phase- and contrast-invariant properties of neurons as well as increasing receptive field sizes along the visual hierarchy~\cite{riesenhuber1999hierarchical,finn2007emergence}. Position and scale invariance also exist in the responses of neurons in area V4~\cite{rust2010selectivity}, but these invariances considerably increase as visual information propagates to neurons in inferior temporal (IT) cortex~\cite{rust2010selectivity, zoccolan2005multiple,zoccolan2007trade,brincat2004underlying,hung2005fast}, where responses are highly consistent when an identical object varies across different dimensions~\cite{murty2015dynamics,cadieu2013neural,yamins2013hierarchical,cadieu2014deep}. In addition, IT cortex is the only area in the ventral stream which encodes three-dimensional transformations through view specific~\cite{logothetis1994view,logothetis1995shape} and view invariant~\cite{perrett1991viewer,booth1998view} responses.

 Inspired by these findings, several early computational models~\cite{Fukushima1980,LeCun1998, Riesenhuber1999, Serre2007.PAMI,Masquelier2007,Lee2009} were proposed. These models mimic feed-forward processing in the ventral visual stream as it is believed that the first feed-forward flow of information, $\sim150$~{\it ms} post-stimulus onset, is usually sufficient for object recognition~\cite{ thorpe1996speed, liu2009timing,freiwald2010functional,anselmi2014unsupervised,hung2005fast}. However, the performance of these models in object recognition was significantly poor comparing to that of humans in the presence of large viewpoint variations~\cite{ghodrati2014feedforward,pinto2011comparing,Pinto2008}.

The second generation of these feed-forward models are called deep convolutional neural networks (DCNNs). DCNNs involve many layers (say 8 and above) and millions of free parameters, usually tuned through extensive supervised learning. These networks have achieved outstanding accuracy on object and scene categorization on highly challenging image databases~\cite{lecun2015deep,krizhevsky2012imagenet,zhou2014learning}. Moreover, it has been shown that DCNNs can tolerate a high degree of variations in object images and even achieve close-to-human performance~\cite{khaligh2014deep,kheradpisheh2015deep,cadieu2014deep}. 
However, despite extensive research, it is still unclear how different types of variations in object images are treated by DCNNs. These networks are position-invariant by design (thanks to weight sharing), but other sorts of invariances must be acquired through training, and the resulting invariances have not been systematically quantified.

In humans, early behavioral studies~\cite{bricolo1993rotation,dill1997translation} showed that we can robustly recognize objects despite considerable changes in scale, position, and illumination; however, the accuracy drops if the objects are rotated in depth. Yet these studies used somewhat simple stimuli (respectively paperclips and combinations of geons). It remains largely unclear how different kinds of variation on more realistic object images, individually and combined with each other, affect the performance of humans, and if they affect the performance of DCNNs similarly.

Here, we address these questions through a set of behavioral and computational experiments in human subjects and DCNNs to test their ability in categorizing object images that were transformed across different dimensions. We generated naturalistic object images of four categories: cars, ships, motorcycles, and animals. Each object carefully varied across either one dimension or a combination of dimensions, among scale, position, in-depth and in-plane rotations. All 2D images were rendered from 3D planes. The effects of variations across single dimension and compound dimensions on recognition performance of humans and two powerful DCNNs~\cite{krizhevsky2012imagenet,simonyan2014very} were compared in a systematic way, using the same images. 

Our results indicate that human subjects can tolerate a high degree of variation with remarkably high accuracy and very short response time. The accuracy and reaction time were, however, significantly dependent on the type of object variation, with rotation in-depth as the most difficult dimension. This finding does not argue in favor of three-dimensional object representation theories, but suggests that object recognition is mainly based on two-dimensional template matching. Interestingly, the results of deep neural networks were highly correlated with those of humans as they could well mimic human behavior when facing variations across different dimensions. This suggests that humans have difficulty to handle those variations that are also computationally more complicated to overcome. More specifically, variations in some dimensions, such as in-depth rotation and scale, that change the amount or the content of input visual information, make the object recognition more difficult for both humans and deep networks.
\section{Material and methods}

\subsection{Image generation}
We generated object images of four different categories: car, animal, ship, and motorcycle. Object images varied across four dimensions: scale, position (horizontal and vertical), in-plane and in-depth rotations. Depending on the type of experiment, the number of dimensions that the objects varied across were determined (see following sections). All two-dimensional object images were rendered from three-dimensional models. There were on average 16 different three-dimensional example models per object category (car: 16, animal: 15, ship: 18, and motorcycle: 16). 

To generate a two-dimensional object image, first, a set of random values were sampled from uniform distributions. Each value determined the degree of variation across one dimension (e.g., size). These values were then simultaneously applied to a three-dimensional object model. This resulted in a three-dimensional object images that has varied across different dimensions. Finally, a two-dimensional image was generated by taking a snapshot from the transformed three-dimensional model. Object images were generated with four levels of difficulty by carefully controlling the amplitude of variations across four levels, from no variation (level 0, where changes in all dimensions were very small: $\Delta_{Sc} = \pm 1\%$, $\Delta_{Po}=\pm 1\%$, $\Delta_{RD} = \pm 1^{\circ}$, and $\Delta_{RP}=\pm 1^{\circ}$; each subscript refers to a dimension: Sc: Scale, Po: Position, RD: in-depth rotation, RP: in-plane rotation; and $\Delta$ is the amplitude of variations) to high variation (level 3: $\Delta_{Sc} = \pm 60\%$, $\Delta_{Po}=\pm 60\%$,  $\Delta_{RD} = \pm 90^{\circ}$, and $\Delta_{RP}=\pm 90^{\circ}$). To control the degree of variation in each level, we limited the range of random sampling to a specific upper and lower bounds. 

\begin{figure*}[!htb]
\centering
\includegraphics[scale=0.105]{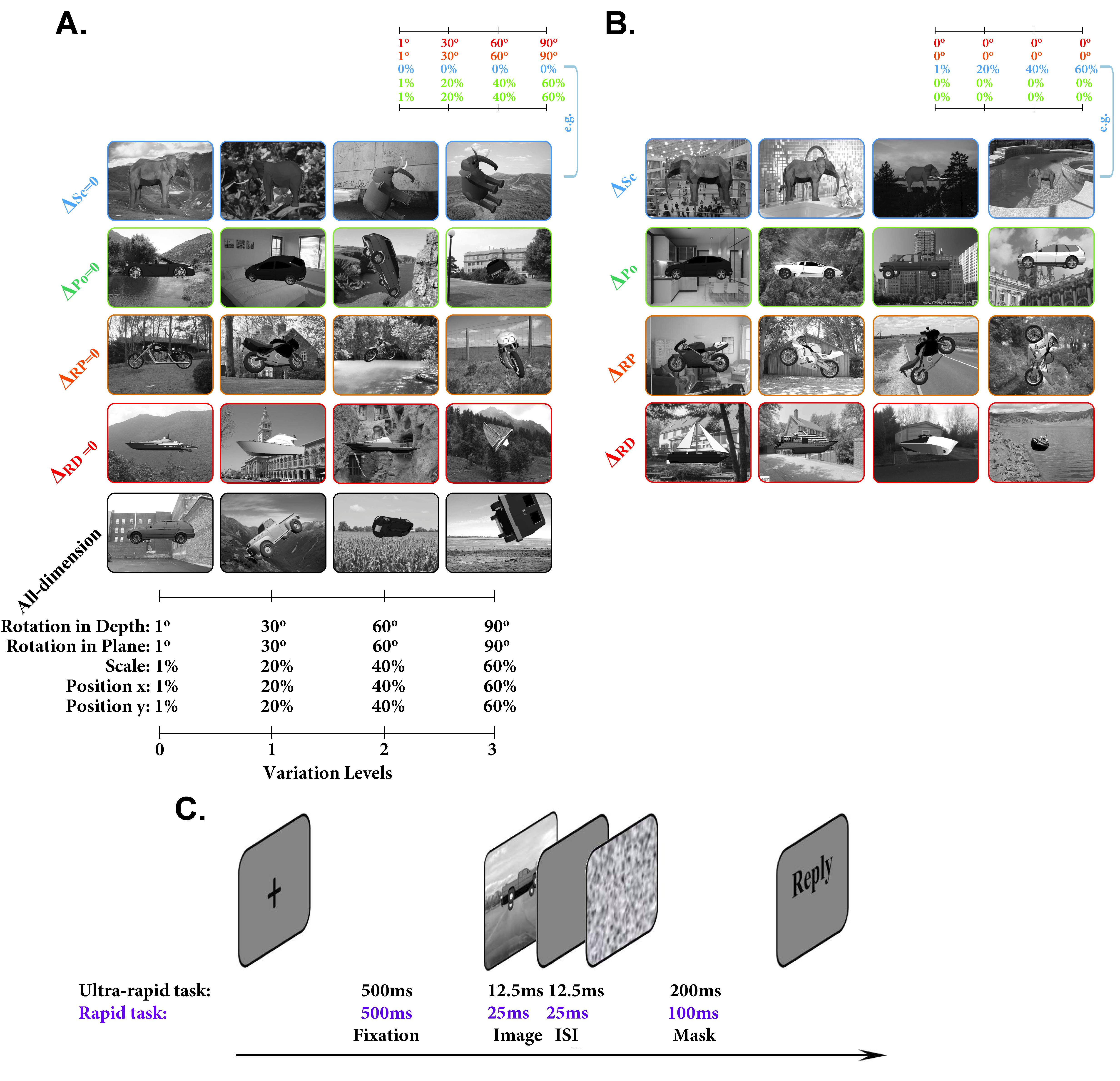}
\caption{\textbf{Image databases and the paradigm of the psychophysical experiment.} A. Sample object images from all- and three-dimension databases for different categories and variation levels. Each column indicates a variation level and each row refers to a database. First four rows are images from three-dimension database: 1st row) $\Delta_{Sc} = 0$; 2nd row) $\Delta_{Po}=0$; 3rd row) $\Delta_{RD} = 0$; and 4th row) $\Delta_{RP}=0$. The last row shows sample images from all-dimension database. The range of variations across different levels is depicted below the images. Colored frames refer to the type of database (this color code is the same throughout the paper). B. Sample images from the one-dimension databases. Each row corresponds to one type of database: 1st row) $\Delta_{Sc}$, \textit{Scale-only}; 2nd row) $\Delta_{Po}$, \textit{Position-only}; 3rd row)  $\Delta_{RP}$, \textit{In-plane-only}; 4th row) $\Delta_{RD}$, \textit{In-depth-only}. The range of variation in each level is the same as A. C. Psychophysical experiment for rapid and ultra-rapid object categorization (see Materials and Methods).}
\label{figure_1}
\end{figure*}

Figure~\ref{figure_1} shows several sample images and the range of variations across four levels. The size of two-dimensional images was $400\times 300$ pixels (width$ \times $height). All images were initially generated on uniform gray background. Moreover, identical object images on natural backgrounds were generated for some experiments. This was done by superimposing object images on randomly selected natural backgrounds from a large pool. Our natural image database contained 3,907 images which consisted of a wide variety of indoor, outdoor, man-made, and natural scenes.

\subsubsection{Different image databases}
To test humans and DCNNs in invariant object recognition tasks, we generated three different image databases:                  

\begin{itemize}
\item \textbf{All-dimension:}  In this database, objects varied across all dimensions, as described earlier (i.e., scale, position, in-depth and in-plane rotations). Object images were generated across four levels in terms of variation amplitude (Level 0-3, Figure~\ref{figure_1}.A). 

\item \textbf{Three-dimension:} The image generation procedure in this database was similar to  all-dimension database, but object images  varied across a combination of three dimensions only, while the 4th dimension was fixed. For example, objects' size were fixed across all variation levels while other dimensions varied (i.e., position, in-depth, and in-plane rotations). This provided us with four databases:  1) $\Delta_{Sc} = 0$: object's size were fixed to a reference size across variation levels; 2) $\Delta_{Po}=0$: objects' position were fixed at the center of the image; 3) $\Delta_{RD} = 0$: objects were not rotated in depth (three-dimensional transformation) across variation levels; 4) $\Delta_{RP}=0$: objects were not rotated in plane (see Figure~\ref{figure_1}.A). 

\item \textbf{One-dimension:} Object images in this database  varied across only one dimension (e.g., size), meaning that the variations across other dimensions were fixed to reference values. Thus, we generated four databases: 1) $\Delta_{Sc}$: only the scale of objects varied; 2) $\Delta_{Po}$: only the position of objects across vertical and horizontal axes varied; 3) $\Delta_{RP}$: objects were only rotated in plane; 4) $\Delta_{RD}$: objects were only rotated in depth. Figure~\ref{figure_1}.B shows several sample images from these databases. 
\end{itemize}

\subsection{Human psychophysical experiments}
We evaluated the performance of human subjects in invariant object recognition through different experiments and using different image databases. In total, the data of 89 subjects (aged between 23-31, mean = 22, 39 female and 50 male) were recorded. Subjects had normal or corrected-to-normal vision. In most experiments, data of 16-20 sessions were recorded  (two experiments were ran for 5 sessions); therefore, some subjects completed all experiments and others only participated in some experiments. All subjects voluntarily participated to the experiments and gave their written consent prior to participation. Our research adhered to the tenets of the Declaration of Helsinki and all experimental procedures were approved by the ethic committee of the University of Tehran.

Images were presented on a 17" CRT monitor (LG T710BH CRT; refresh rate 80Hz, resolution 1280$ \times $1024 pixels) connected to a PC equipped with an NVIDIA GeForce GTX 650 graphic card. We used MATLAB (www.mathworks.com) with psychophysics toolbox~\cite{brainard1997psychophysics, pelli1997videotoolbox} (http://psychtoolbox.org) to present images. Subjects had a viewing distance of 60 {\it cm} and each image covered $\sim 10 \times 11$ degrees of visual angle. 

Details of each experiment are explained in the following sections. Generally, we used rapid image presentation paradigm with mask to only account  for the feed-forward processing in the ventral visual pathway (see Figure~\ref{figure_1}). Each trial was started with a black fixation cross, presented at the center of the screen for 500~{\it ms}, followed by a randomly selected image from the database that was presented for either one or two frames depending on the experiment type (see the description for each experiment). Afterwards, a gray blank screen was presented as inter-stimulus interval (ISI). Finally, a $1/f$ noise mask image was presented. The timing of the image presentation, ISI, and mask depended on the experiment type (see the following sections). Subjects' task was to categorize the presented object images. They were instructed to  respond as fast and accurate as they could by pressing a key on computer keyboard (each key was labeled with a category name). The next trial was started after the key press and there was a random time delay before the start of the next trial. We recorded both subjects' reaction times (reported in Supplementary Information) and accuracies.  Each experiment was divided into a number of blocks and subjects could rest between blocks. Subjects performed some practice trials prior to the main experiment. Images in the practice trials were not presented in the main experiments. In the next sections, we describe the details of each experiment. 

\subsubsection{Rapid invariant object categorization}\label{Multi-class-psycho}
In these experiments, subjects categorized rapidly presented images from four object categories (see Figure~\ref{figure_1}). Each trial started with a fixation cross presented at the center of the screen for 500~{\it ms}. An image was then randomly selected from the pool and was presented for 25~{\it ms} (2 frames of 80 Hz monitor) followed by a gray blank screen for 25~{\it ms} (ISI). Immediately after the blank screen, a $1/f$ noise mask image was presented for 100~{\it ms}. Subjects were asked to rapidly and accurately press one of the four keys, labeled on keyboard, to indicate which object category was presented. The next trial started after a key press with a random time delay ($2 \pm 0.5$ second). This experiment was performed in two types that are explained as following:  

\begin{itemize}
 \item \textbf{Using the all-dimension database:} We used  all-dimension database where object images varied across all dimensions (i.e., scale, position, in-depth and in-plane rotations). In each session, subjects were presented with 320 images: 4 categories $\times$ 4 levels $\times$ 20 images per category. Images were presented into two blocks of 160 images. For each background condition (i.e., uniform and natural backgrounds), we recorded the data of 16 different sessions.                

\item \textbf{ Using three-dimension databases:} In this experiment, we used the three-dimension databases. Using these databases, we could measure how excluding variations across one dimension can affect human performance in invariant object recognition: if the fixed dimension is more difficult than the others, subjects would be able to categorize the object more accurately and within shorter time than in the case where the fixed dimension is much easier. We presented subjects with 960 images: 4 categories $\times$ 4 levels $\times$ 4 conditions ($\Delta_{Sc} = 0$, $\Delta_{Po} = 0$, $\Delta_{RP} = 0$, and $\Delta_{RD} = 0$) $\times$ 15 images per category. Images were presented in four blocks of 240 images.  Note that we inter-mixed images of all conditions in each session; therefore, subjects were unaware of the type of variations. We recorded the data of 20 sessions for the case of objects on natural backgrounds and 17 sessions for objects on uniform background.
\end{itemize}

To have more accurate reaction times, we also performed two-category (car versus animal) rapid invariant object categorization tasks with similar experimental settings. The details of these two-category experiments and their results are presented in Supplementary Information. 

\subsubsection{Ultra-rapid invariant object categorization} \label{Ultra-Rapid}
To assess whether the experimental design (presentation time and variation conditions) could affect our results and interpretations, we ran two ultra-rapid invariant object categorization tasks, using {\it three-dimension} and one-dimension databases. In each trial, we presented a fixation cross for 500~{\it ms}. Then, an image was randomly selected from the pool and presented to the subject for 12.5~{\it ms} (1 frame at 80~Hz monitor). The image was then followed by a blank screen for 12.5~{\it ms}. Finally, a noise mask was presented for 200~{\it ms}. Subjects had to accurately and rapidly press one of the four keys, labeled on the keyboard, to declare their responses. The next trial  started after a key press with a random time delay ($2 \pm 0.5$ second). As mentioned above, this experiment was performed in two types that are explained as following:  

\begin{itemize}
\item \textbf{Using three-dimension database:} We recorded the data of five human subjects. Object images were selected from natural background three-dimension database. Images are identical to those of three-dimension rapid presentation experiment described in previous section. But, here, images were presented for 12.5~{\it ms} followed by~12.5~{\it ms} blank and then 200~{\it ms} noise mask.

\item \textbf{Using one-dimension database:} In this experiment, we used natural background one-dimension database to evaluate the effect of variations across individual dimensions on human performance. Subjects were presented with 960 images: 4 categories $\times$  4 levels $\times$  4 conditions ($\Delta_{Sc}$,$\Delta_{Po}$,$\Delta_{RP}$,$\Delta_{RD}$) $\times$  15 images per category. The experiment was divided into four blocks of 240 images. We collected the data of five sessions. Note that we only used objects on natural backgrounds because this task was easier compared to previous experiments; therefore, categorizing objects on uniform background would be very easy. For the same reason, we did not used the one-dimension database in the rapid task.
\end{itemize}

\subsection{Behavioral data analysis}
We calculated the accuracy of subjects in each experiment as the ratio of correct responses (i.e., Accuracy \% = 100 $\times$ Number of correct trials / Total number of trials). The accuracies of all subjects were calculated and the average and standard deviation were reported. We also calculated confusion matrices for different conditions of rapid invariant object categorization experiments, which are presented in Supplementary Information. A confusion matrix allowed us to determine which categories were more miscategorized and how categorization errors were distributed across different categories. To calculate the human confusion matrix for each variation condition, we averaged the confusion matrices of all human subjects.

We also analyzed subjects' reaction times in different experiments which are provided in Supplementary Information. In the two-category experiment, first, we removed reaction times longer than 1200~{\it ms} (only 7.8\% of reaction times were removed across all experiments and subjects). We then compared the reaction times in different experimental conditions. The reported results are the mean and standard deviation reaction times. In four-category experiments, we removed reaction times longer than 1500~{\it ms} because in these tasks it could take longer time to press a key (only 8.7\% of reaction times were removed across all experiments and subjects). Although the reaction times in four-category experiments might be a bit unreliable as subjects had to select one key out of four keys, they provided us with clues about the effect of variations across different dimensions on humans' response time.

\subsection{Deep convolutional neural networks (DCNNs)}
DCNNs are a combination of deep learning~\cite{schmidhuber2015deep} and convolutional neural networks~\cite{LeCun1998}.  DCNNs use  a hierarchy of several consecutive feature detector layers. The complexity of features increases along the hierarchy. Neurons/units in higher convolutional layers are selective to complex objects or object parts. Convolution is the main process in each layer that is generally followed by complementary operations such as pooling and output normalization. Recent deep networks, which have exploited supervised gradient descend based learning algorithms, have achieved remarkable performances in recognizing extensively large and difficult object databases such as Imagenet~\cite{lecun2015deep,schmidhuber2015deep}. Here, we evaluated the performance of two of the most powerful DCNNs proposed by Krizhevsky et al.~\cite{krizhevsky2012imagenet} and Simonyan and Zisserman~\cite{simonyan2014very} in invariant object recognition. More explanations about these networks are provided as following:

\begin{itemize}
\item \textbf{Krizhevsky et al. (2012):} This model achieved an impressive performance in categorizing object images for Imagenet database and significantly defeated other competitors in the ILSVRC-2012 competition~\cite{krizhevsky2012imagenet}.  Briefly, the model contains five convolutional (feature detector) and three fully connected (classification) layers. The model uses Rectified Linear Units (ReLUs) as the activation function of neurons. This significantly sped up the learning phase. The max-pooling operation is performed in the first, second, and fifth convolutional layers. The model is trained using a stochastic gradient descent algorithm. This network has about 60 millions free parameters. To avoid overfitting during the learning procedure, some data augmentation techniques (enlarging the training set) and the dropout technique (in the first two fully-connected layers) were applied. Here, we used the pre-trained version of this model (on the Imagenet database) which is publicly available at \url{http://caffe.berkeleyvision.org} by Jia et.~al~\cite{jia2014caffe}.

\item \textbf{Very Deep (2014):}An important aspect of DCNNs is the number of internal layers, which influences their final performance. Simonyan and Zisserman~\cite{simonyan2014very}  studied the impact of the network depth  by implementing  deep convolutional networks with 11, 13, 16, and 19 layers. For this purpose, they used very small convolution filters in all layers, and steadily increased the depth of the network by adding more convolutional layers. Their results showed that the recognition accuracy increases by adding more layers and the 19-layer model significantly outperformed other DCNNs. Here, we used the 19-layer model which is freely available at \url{http://www.robots.ox.ac.uk/~vgg/research/very_deep/}. 
\end{itemize}

\subsection{Evaluation of DCNNs}
We evaluated the categorization accuracy of deep networks on natural background three- and one-dimension tasks. To this end, we first randomly selected 600 images from  each object category, variation level, and variation condition (three- or one-dimension). Hence, we used 8 different image databases (4 variation levels $\times$ 2 variation conditions), each of which consisted of 2500 images (4 categories $\times$ 600 images). To compute the accuracy of each DCNN for given variation  condition and  level,  we randomly selected two subsets of 1200 training (300 images per category) and 600 testing images (150 images per category) from the corresponding image database. We then fed the DCNN with the training and testing images and calculated the corresponding feature vectors of the last convolutional layer. Afterwards, we used these feature vectors to train the classifier and compute the categorization accuracy. Here we used a linear SVM classifier (libSVM implementation~\cite{CC01a}, \url{www.csie.ntu.edu.tw/~cjlin/libsvm}) with  optimized regularization parameters. This procedure was repeated for 15 times (with different randomly selected training and testing sets) and the average and standard deviation of the accuracy were computed. This procedure was done for both DCNNs and all variation conditions and levels. 

To visualize the similarity between DCNNs and humans, we performed a Multidimensional Scaling (MDS) analysis across the variation levels of the three-dimension task.  For each human subject or DCNN, we put together its accuracies over different variation conditions in a vector. Then we plotted the 2D MDS map based on the cosine similarities (distances) between these vectors. 

\begin{figure}[!tb]
\centering
\includegraphics[scale=1.8]{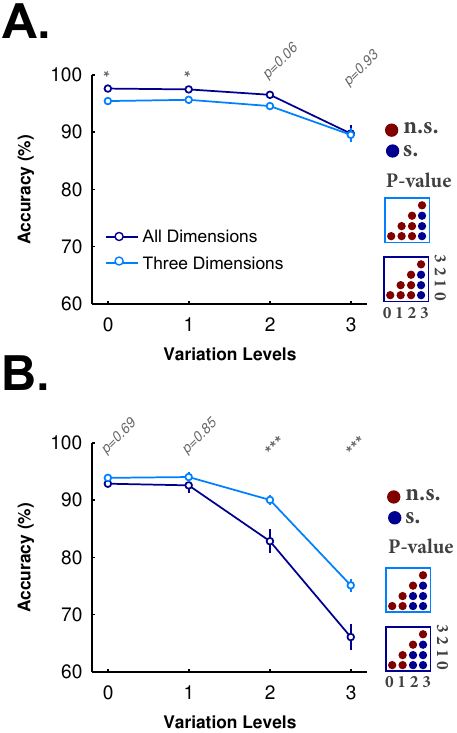}
\caption{\textbf{Accuracy of subjects in rapid invariant object categorization task.} A. The accuracy of subjects in categorization of four object categories, when objects had uniform backgrounds. The dark, blue curve shows the accuracy when objects varied in all dimensions and the light, blue curve demonstrates the accuracy when objects varied in three dimensions. Error bars are the standard deviation (STD). P values, depicted at the top of curves, show whether the accuracy between all- and three-dimension experiment are significantly different (Wilcoxon rank sum test). Color-coded matrices, at the right, show whether changes in accuracy across levels statistically significant (Wilcoxon rank sum test; each matrix corresponds to one curve; see color of the frame). B. Categorization accuracy when objects had natural backgrounds.}
\label{figure_2}
\end{figure}

\section{Results}
We ran different experiments in which subjects and DCNNs categorized object images varied across several dimensions (i.e., scale, position, in-plane and in-depth rotations, background). We measured the accuracies and reaction times of human subjects in different rapid and ultra-rapid invariant object categorization tasks, and the effect of variations across different dimensions on human performance was evaluated. The human accuracy was then compared with the accuracy of two well-known deep networks~\cite{krizhevsky2012imagenet,simonyan2014very} performing the same tasks as humans. We first report human results in different experiments and then compare them with the results of deep networks.

\subsection{Human performance is dependent on the type of object variation}
In these experiments, subjects were asked to accurately and quickly categorize rapidly presented object images of four categories (animal, car, motorcycle, and ship) appeared in uniform and natural backgrounds (see section~\ref{Multi-class-psycho}).

\begin{figure*}[!htb]
\centering
\includegraphics[scale=1.4]{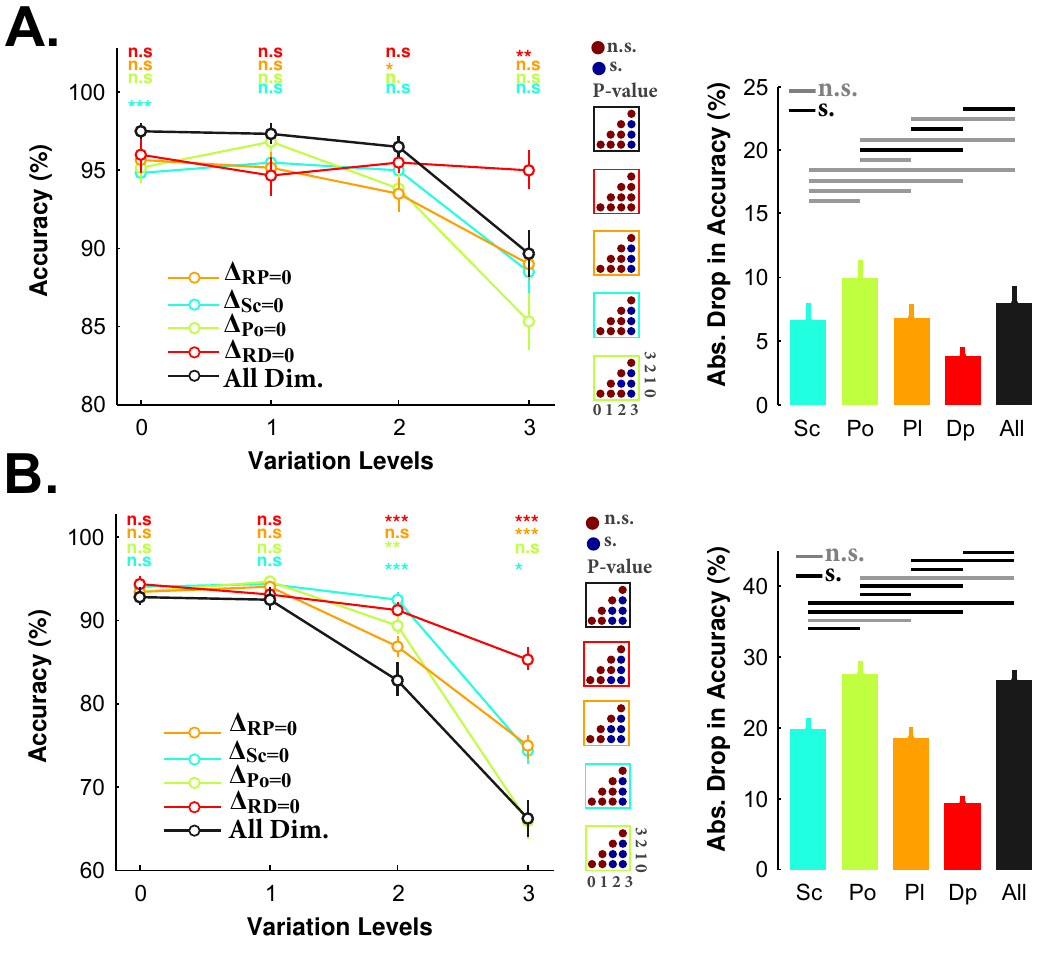}
\caption{\textbf{Accuracy of subjects in rapid invariant object categorization task for all-dimension and different three-dimension conditions.} A. Accuracies for uniform background experiments. Left, The accuracy of subjects in categorization of four object categories (i.e., car, animal, ship, motorcycle). Each curve corresponds to one condition: $\Delta_{Sc}=0 $, $ \Delta_{Po}=0 $, $ \Delta_{RP}=0 $, $ \Delta_{RD}=0 $ (as specified with different colors). Error bars are the standard deviation (STD). P values, depicted at the top of curves, show whether the accuracy between all-dimension and other three-dimension conditions are significantly different (Wilcoxon rank sum test). Color-coded matrices, at the right, show whether changes in accuracy across levels are statistically significant (e.g., accuracy drop is significant from one level to the other; Wilcoxon rank sum test; each matrix corresponds to one curve; see color of the frame). Right, absolute accuracy drop between level 0 and level 3 (mean+/-STD). The horizontal lines at the top of bar plot shows whether the differences are significant (gray line: insignificant, black line: significant). B. Accuracies for natural backgrounds experiments. Figure conventions are similar to A.}
\label{figure_3}
\end{figure*}

Figures~\ref{figure_2}.A and~\ref{figure_2}.B provide the average accuracy of subjects over different variation levels in all- and three-dimension conditions while objects had uniform and natural backgrounds, respectively. Figure~\ref{figure_2}.A shows that there is a small and negligible difference between the categorization accuracies in all- and three-dimension conditions with objects on uniform background. Also, for both experimental conditions, the categorization errors significantly increased at high variation levels (see the color-coded matrices in the right side of Figure~\ref{figure_2}.A). Despite the small, but significant, accuracy drop, this data shows that humans can robustly categorize object images when they have uniform background even at the highest variation levels (average accuracy above 90\%). In addition, the reaction times in all- and three-dimension experiments were not significantly different (Figure~S9.A).

Conversely, in the case of objects on natural backgrounds (Figure~\ref{figure_2}.B), the categorization accuracies in both experimental conditions substantially decreased as the variation level was increased (see the color-coded matrices in the right side of Figure~\ref{figure_2}.B; Wilcoxon rank sum test), pointing out the difficulty of invariant object recognition in clutter. Moreover, in contrast to the uniform background experiments, there is a large significant difference between the accuracies in all- and three-dimension experiments (see p values depicted at the top of Figure~\ref{figure_2}.B; Wilcoxon rank sum test). Overall, it is evident that excluding one dimension can considerably reduce the difficulty of the task. A similar trend can be seen in the reaction times (see Figure~S9.B), where the reaction times in both conditions significantly increased as the variation level increased.

We then broke the trials into different conditions and calculated the mean accuracy in each condition (i.e., $ \Delta_{Sc}=0 $, $ \Delta_{Po}=0 $, $ \Delta_{RP}=0 $, $ \Delta_{RD}=0 $ ). Figure~\ref{figure_3}.A demonstrates the accuracies in all and three-dimension conditions, for the case of objects on uniform background. As seen, there is a small difference in the accuracies of different conditions at low and intermediate variation levels (level 0-2). However, at the highest variation level, the accuracy in $ \Delta_{RD}=0 $ (red curve) is significantly higher than the other conditions, suggesting that excluding in-depth rotation made the task very easy despite variations across other dimensions. Note that in $ \Delta_{RD}=0 $ the accuracy curve is virtually flat across levels with average of $\sim$95\%. Interestingly, the accuracies were not significantly different between all-dimension experiment and $ \Delta_{Po}=0 $, $ \Delta_{Sc}=0 $, and $ \Delta_{RP}=0 $. This confirms that much of the task difficulty arises from in-depth rotation, although other dimensions have some weaker effects (e.g., scale, and rotation in-plane). This is also reflected in the bar plot in Figure~\ref{figure_3}.A as the absolute accuracy drop in $ \Delta_{RD}=0 $  is lower than 5\%, while it is above 10\% in $ \Delta_{Po}=0 $. It is also clear that humans had the maximum errors in $ \Delta_{Po}=0 $ condition, suggesting that removing position variation did not considerably affect the task difficulty (i.e., position is the easiest dimension).

The reaction times were compatible with the accuracy results (see Figure~S10.A), where at the highest variation level, the human reaction times  in $ \Delta_{Sc}=0 $, $ \Delta_{Po}=0 $, and $ \Delta_{RP}=0 $ significantly increased, while it did not significantly change in $ \Delta_{RD}=0 $. In other words, when objects were not rotated in-depth, humans could quickly and accurately categorize them. 

In a separate experiment, subjects performed similar task while objects had natural backgrounds. Results show that there were small differences between the accuracies in all-dimension and three-dimension conditions at the first two variation levels (Figure~\ref{figure_3}.B). This suggests that human subjects could easily categorize object images on natural backgrounds while objects had small and intermediate degree of variations. However, accuracies became significantly different as the variation level increased (e.g., levels 2 and 3 see color-coded matrices in Figure \ref{figure_3}.B). As shown in Figure~\ref{figure_3}.B, there is about 20\% accuracy difference between $ \Delta_{RD}=0 $ and all-dimension condition  at the most difficult level, confirming that the rotation in depth is a very difficult dimension. The bar plot in Figure~\ref{figure_3}.B, shows that the highest accuracy drop, between levels 0 and 3, belonged to $ \Delta_{Po}=0 $ and all-dimension conditions while the lowest drop was observed in $\Delta_{RD}=0$. In addition, the accuracies in  $\Delta_{Sc}=0 $ and $\Delta_{RP}=0 $ fall somewhere between $\Delta_{Po}=0 $ and $\Delta_{RD}=0$, indicating that scale variations and in-plane rotation imposed more difficulty than variations in position; however, they were easier than rotation in depth. This is also evident in the accuracy drop. 

Different objects have different three-dimensional properties; so, the categorization performance might be affected by these properties. In this case, one object category might bias the performance of humans in different variation conditions. To address this question, we broke the trials into different categories and calculated the accuracies (Figure~S5) and reaction times (Figures~S10.B and S11.B) for all variation and background conditions. The results indicated that although the categorization accuracy and reaction time may differ between categories,  the order of the difficulty of different variation conditions are consistent in all categories. That is in-depth rotation and position transformation are respectively the most difficult and easy variations to process. We also calculated the confusion matrix of humans for each variation condition and level, to have a closer look at error rate and miscategorization across categories. The confusion matrices for uniform and natural background experiments are presented in Figure~S6.

Analyses so far have provided information about the dependence of human accuracy and reaction time on the variations across different dimensions. However, one may ask how these results can be influenced by low-level image properties such as luminance and contrast. To address this, we computed the correlation between low-level image statistics (contrast and luminance) and the performance of human subjects. The results show that neither luminance (Figure~S7) nor contrast (Figure~S8) could explain human accuracy and reaction time in our invariant object recognition tasks.

We also performed similar two-category rapid tasks and their results are provided int Supplementary Information(Figure~S1-S4). Interestingly, the results of two-category experiments are consistent with the four-category tasks, indicating that our results are robust to number of categories.

\begin{figure*}[!htb]
\centering
\includegraphics[scale=1.25]{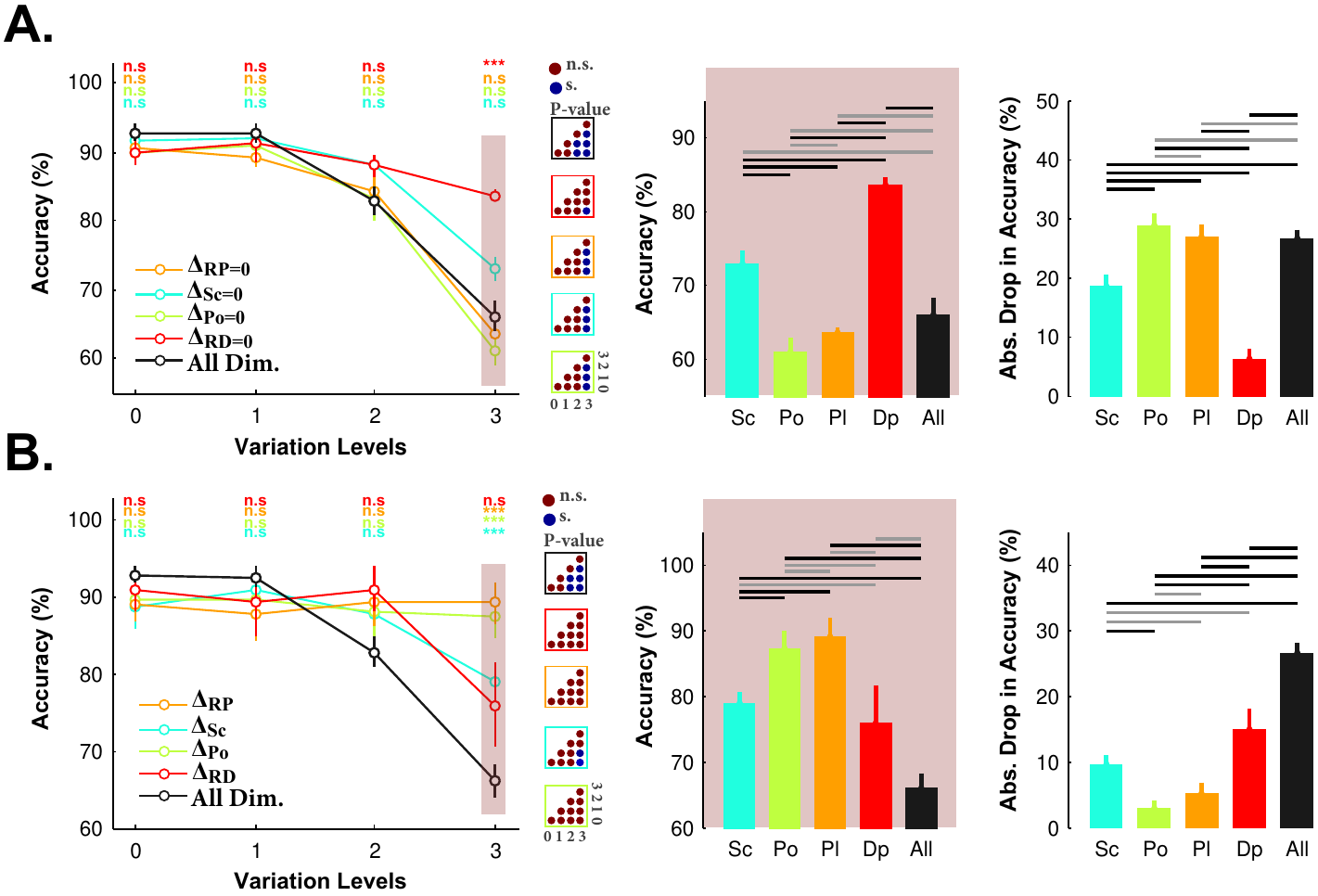}
\caption{\textbf{Accuracy of human subjects in ultra-rapid invariant object categorization task for three- and one-dimension conditions, when objects had natural backgrounds.} A. Left, the accuracy of human subjects in three-dimension experiments. Each curve corresponds to one condition: $\Delta_{Sc}=0 $, $ \Delta_{Po}=0 $, $ \Delta_{RP}=0 $, $ \Delta_{RD}=0 $ (as specified with different colors). Error bars are the standard deviation (STD). Color-coded matrices, on the right show whether changes in accuracy across levels in each condition are statistically significant (e.g., accuracy drop is significant from one level to the other; Wilcoxon rank sum test; each matrix corresponds to one curve; see color of the frame). Note that the results of the average and STD of 5 subjects. Middle, categorization accuracy in level 3 in different three-dimension conditions (each bar corresponds to a condition). The horizontal lines on top of the bar plot shows whether the differences are significant (gray line: insignificant, black line: significant). Right, absolute accuracy drop between level 0 and level 3 (mean+/-STD). Each bar, with specific color, corresponds to one condition. B. Similar to part A, where the plots present the results in one-dimension experiments.}
\label{figure_4}
\end{figure*}

\subsection{Human performance is independent of experimental setup}
Although the effect of variations across different dimensions of an object on subjects' performance was quite robust, we designed two other experiments to investigate how decreasing the presentation time would affect our results. Therefore, we reduced the image presentation time and the following blank screen from 25~{\it ms} to 12.5~{\it ms} (ultra-rapid object presentation). We also increased the time of the subsequent noise mask from 100~{\it ms} to 200~{\it ms}. In the first experiment, we repeated the natural background three-dimension categorization task with the ultra-rapid settings. We did not run uniform background condition because our results showed that this task would be easy and some ceiling effects may mask differences between conditions. For the second experiment, we studied the effect of each individual dimension (e.g., scale only) on the accuracy and reaction time of subjects. In the following, we report the results of these two experiments.

\subsubsection{Shorter presentation time does not affect human performance}
Figure~\ref{figure_4}.A illustrates the results of the ultra-rapid object categorization task in three-dimension conditions with objects on natural backgrounds. Comparing the results in rapid (see Figure~\ref{figure_3}.B) and ultra-rapid experiments (see Figure~\ref{figure_4}.A, the left plot) indicates that there is no considerable difference between the accuracies in these two experiments. This shows the ability of human visual system to extract sufficient information for invariant object recognition even under ultra rapid presentation. 

Similar to the rapid experiment, subjects had the highest categorization accuracy in $\Delta_{RD}=0$ condition, even at the most difficult level, with significant difference to other conditions (see the middle plot in Figure~\ref{figure_4}.A). However, there is a significant difference in accuracies ($\sim 10\%$) between $\Delta_{Sc}=0$ and $\Delta_{RP}=0$. In other words, tolerating scale variation seems to be more difficult than in-plane rotation in ultra-rapid presentation task. It suggests that it is easier to recognized a rotated object in plane than a small object. Comparing the accuracies in level 3 indicates that $\Delta_{RD}=0$ and  $\Delta_{Sc}=0$ were the easiest tasks while $\Delta_{Po}=0$ and $\Delta_{RP}=0$ were the most difficult ones. Moreover, although there was no significant difference in reaction times of different conditions (Figure~S12.A), subjects had shorter reaction times in $\Delta_{RD}=0$ at level 3 while the reaction times were longer in $\Delta_{Po}=0$ at this level. 

Overall, the results of ultra-rapid experiment showed that different time setting did not change our initial results about the effect of variations across different dimensions, despite imposing higher task difficulty.
 
\subsubsection{Humans have a consistent behavior in the one-dimension experiment}
In all experiments so far, object images varied across more than one dimension. In this experiment, we evaluated the performance of human subjects in ultra-rapid object categorization task while objects varied across a single dimension. Object images were presented on natural backgrounds. Figure~\ref{figure_4}.B illustrates that the accuracies were higher in $\Delta_{RP}$ and $\Delta_{Po}$ than in $\Delta_{RD}$ and $\Delta_{Sc}$ conditions. Hence, similar to results shown in Figure~\ref{figure_4}.A for three-dimension experiments, variations across position and in-plane rotation were easier to tolerate than in scale and in-depth rotation (again the most difficult). Subjects also had the highest accuracy drop between levels 0 and 3 in $\Delta_{RD}$ and $\Delta_{Sc}$ conditions while the accuracy drop in $\Delta_{RP}$ was significantly lower (bar plots in Figure~\ref{figure_4}.B). 

The reaction times in different conditions are shown in Figure~S12.B. Although the differences were not statistically significant, the absolute increase in reaction time in $\Delta_{Sc}$ and $\Delta_{RD}$ was higher than the other conditions, confirming that these variations needed more processing time (note that the results are average of five subjects, and increasing the number of subjects might lead to  significant differences).

\begin{figure*}[!htb]
\centering
\includegraphics[scale=.85]{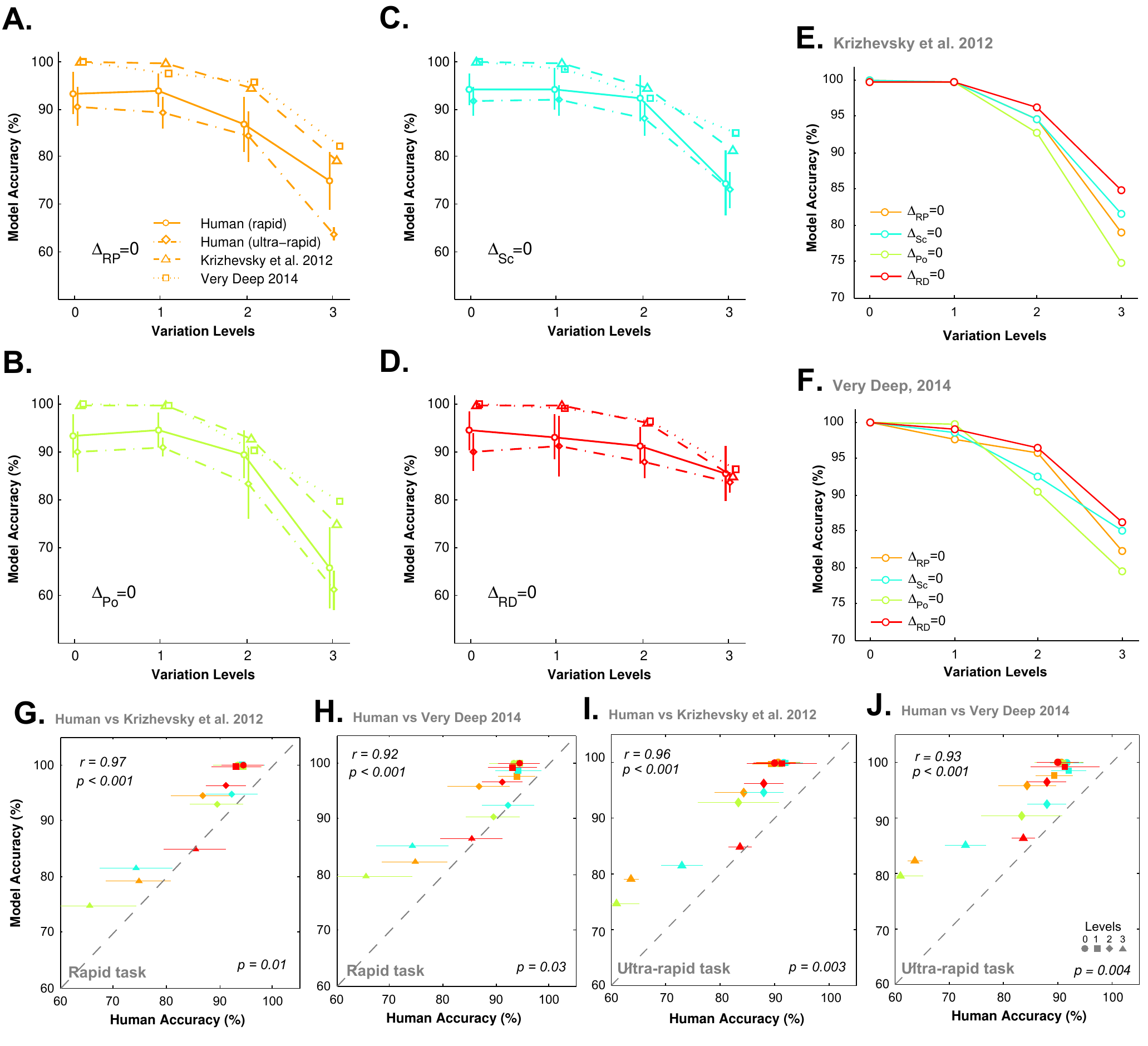}
\caption{\textbf{The accuracy of DCNNs compared to humans in rapid and ultra-rapid three-dimension object categorization tasks.} A-D. The accuracy of Very Deep (dotted line) and Krizhevsky models (dashed line) compared to humans  in categorizing images from three-dimension database while objects had natural background. E. and F. The average accuracy of DCNNs in different conditions. G-H. Scatter plots of human accuracy in rapid three-dimension experiment against the accuracy of DCNNs. I-J. Scatter plot of human accuracy in ultra-rapid three-dimension experiment against the accuracy of DCNNs. Colors show different condition and marker shapes refer to variation levels. The correlation is depicted on the upper-left and the {\it p-value} on lower-right shows whether human and models are significant.}
\label{figure_5}
\end{figure*}

\begin{figure*}[!htb]
\centering
\includegraphics[scale=.88]{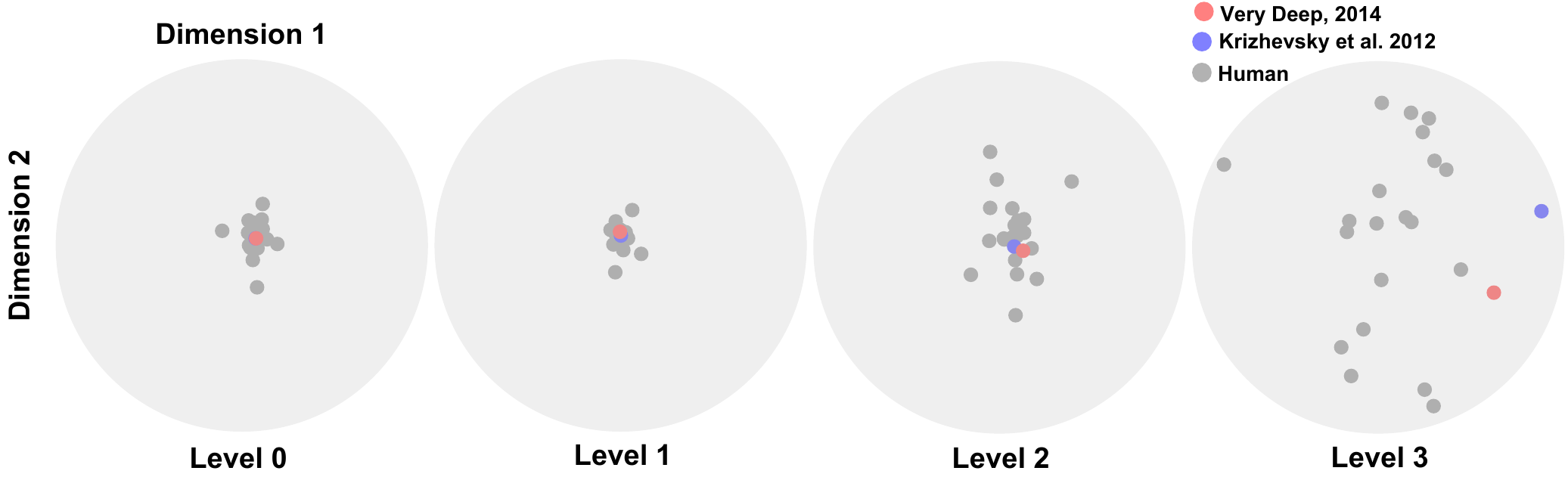}
\caption{\textbf{The similarity between DCNNs and humans.} Scatter plots obtained using multidimensional scaling (MDS). Each plot shows the similarity distances for a variation level. Gray dots illustrate human subjects while red (Very Deep) and blue (Krizhevsky) dots refer to the DCNNs}. 
\label{figure_6}
\end{figure*}

\subsection{DCNNs perform similarly to humans in different experiments}
We examined the performance of two powerful DCNNs on our three- and one-dimension databases with objects on natural backgrounds. We did not use gray background because it would be too easy. The first DCNN was the 8-layer network, introduced by Krizhevsky et~al.~\cite{krizhevsky2012imagenet}, and the second was a 19-layer network, also known as Very Deep model, proposed by Simonyan and Zisserman~\cite{simonyan2014very}. These networks achieved great performance on Imagenet as one of the most challenging current images databases.

Figures~\ref{figure_5}.A-D compares the accuracies of DCNNs with humans (for both rapid and ultra-rapid experiments) on different conditions of three-dimension database (i.e., $\Delta_{RP}=0$, $\Delta_{Po}=0$, $\Delta_{Sc}=0$, $\Delta_{RD}=0$). Interestingly, the overall trend in accuracies of DCNNs were very similar to humans in different variation conditions of both rapid and ultra-rapid experiments. However, DCNNs outperformed humans in different tasks. Despite significantly higher accuracies of both DCNNs compared to humans, DCNNs accuracies were significantly correlated with those of humans for both rapid (Figure~\ref{figure_5}.G-H) and ultra-rapid (Figure~\ref{figure_5}.I-J) experiments. In other words, deep networks can resemble human object recognition behavior in the face of different types of variation. Hence, if a variation is more difficult (easy) for humans, it is also more difficult (easy) for DCNNs. 

We also compared the accuracy of DCNNs in different experimental conditions (Figure~\ref{figure_5}.E-F). Figure~\ref{figure_5}.E shows that the Krizhevsky network could easily tolerate variations in the first two levels (levels 0 and 1). However, the performance decreased at higher variation levels (levels 2 and 3). At the most difficult level (level 3), the accuracy of DCNNs were highest in $\Delta_{RD}=0$ while this significantly dropped to lowest accuracy in $\Delta_{Po}=0$. Also, accuracies were higher in $\Delta_{Sc}=0$ than $\Delta_{RP}=0$. Similar result was observed for Very Deep model with slightly higher accuracy (Figure~\ref{figure_5}.F).

We performed a MDS analysis to visualize the similarity between the accuracy patterns of DCNNs and all human subjects across variation levels (see Materials and methods). For this analysis, we used the rapid categorization data only (20 subjects), and not the ultra rapid one (5 subjects only, which is not sufficient for MDS). Figure~\ref{figure_6} shows that the similarity between DCNNs and humans is high at the first two variation levels. In other words, there is no difference between humans and DCNNs in low variation levels and DCNNs treat different variations as humans. However, the distances between DCNNs and human subjects increased at the third level and became greater at the last level.  This points to the fact that as the level of variation increases the task becomes difficult for both humans and DCNNs and the difference between them increases. Although DCNNs get further away from humans, it is not much greater than human inter-subject distances. Hence, it can be said that even in higher variation levels DCNNs perform similarly to humans. Moreover, the Very Deep network is closer to humans than the Krizhevsky model. This might be the result of exploiting more layers in Very Deep network which helps it to act more like humans.

\begin{figure*}[!htb]
\centering
\includegraphics[scale=.93]{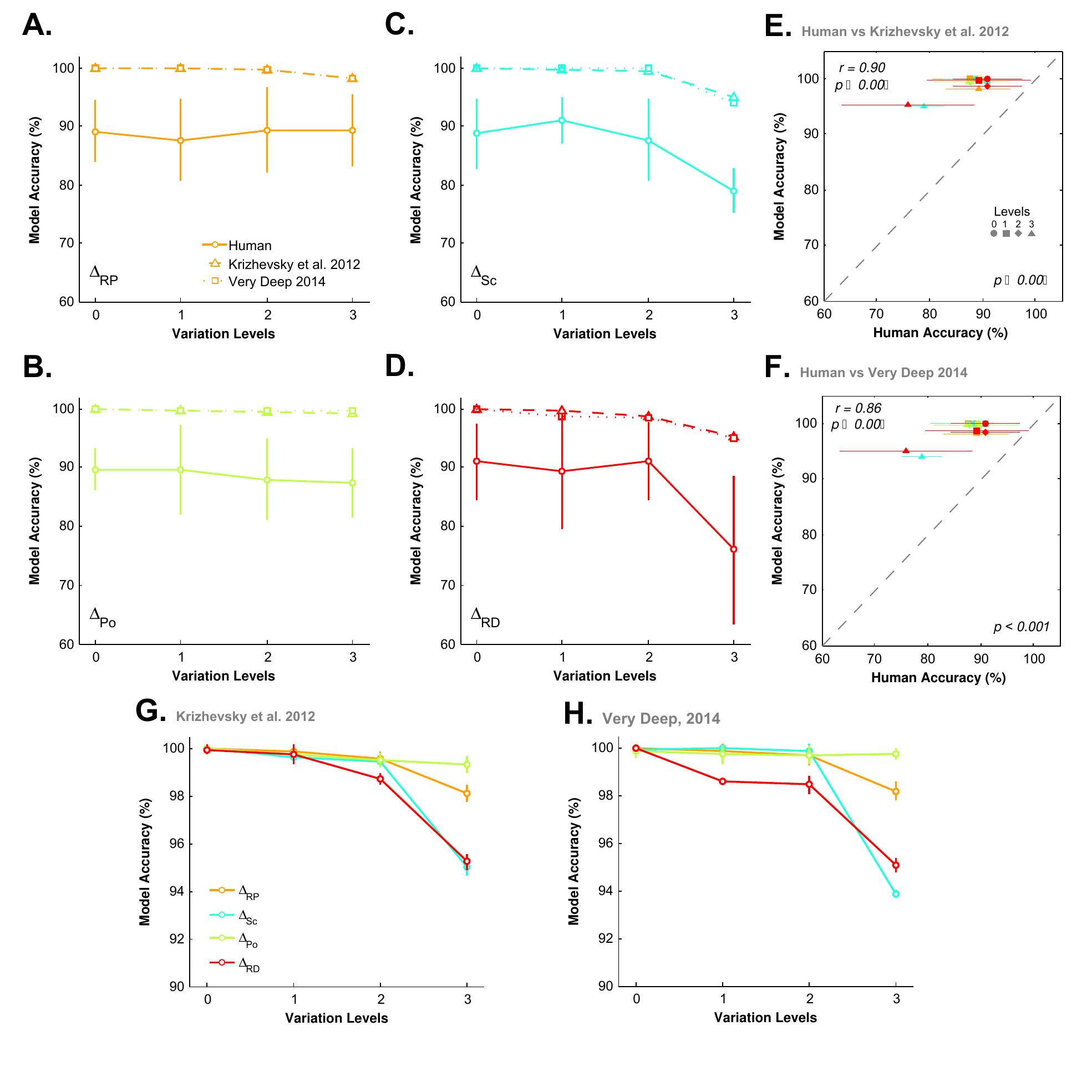}
\caption{\textbf{The accuracy of DCNNs compared to humans in invariant object categorization.} A-D. The accuracy of Very Deep (dotted line) and Krizhevsky models (dashed line) compared to humans (solid line) in categorizing images from one-dimension database while object had natural background. E and F. Scatter plot of human accuracy against the accuracy of DCNNs. Colors show different condition and marker shapes refer to variation levels. The correlation is depicted on the upper-left and the {\it p-value} on lower-right shows whether human and models are significant. G and H. The average accuracy of DCNNs in different condition }
\label{figure_7}
\end{figure*}

To compare DCNNs with humans in the one-dimension experiment, we also evaluated the performance of DCNNs using one-dimension database with natural backgrounds (Figure~\ref{figure_7}). Figure~\ref{figure_7}.A-D illustrates that DCNNs outperformed humans across all conditions and levels. The accuracy of DCNNs was about 100\% at all levels. Despite this difference, we observed a significant correlation between the accuracies of DCNNs and humans (Figure~\ref{figure_7}.E-F), meaning that when a condition was difficult for humans it was also difficult for models.  

To see how the accuracy of DCNNs depends on the dimension of variation, we re-plotted the accuracies of the models in different conditions (Figure~\ref{figure_7}.G-H). It is evident that both DCNNs performed perfectly in $\Delta_{Po}$, which is possibly inherent by their network design (the weight sharing mechanism in DCNNs~\cite{kheradpisheh2015bio}), while they achieved relatively lower accuracies in  $\Delta_{Sc}$ and $\Delta_{RD}$. Interestingly, these results are compatible with humans' accuracy over different variation conditions of one-dimension psychophysics experiment (Figure~\ref{figure_4}), where the accuracies of $\Delta_{Po}$ and $\Delta_{RP}$ were high and almost flat across the levels and the accuracies of $\Delta_{Sc}$ and $\Delta_{RD}$ were lower and significantly dropped in the highest variation level. It can be interpreted that, those variations which change the amount or the content of input visual information, such scaling and in-depth rotation, are much harder to handle (for both humans and DCNNs) than other types of variation such as position transformation and in-plane rotation. 

\section{Discussion}
Although it is well known that the human visual system can invariantly represent and recognize various objects, the underlying mechanisms are still mysterious. Most studies have used object images with very limited variations in different dimensions, presumably to decrease experiment and analysis complexity. Some studies investigated the effect of a few variations on neural and behavioral responses (e.g., scale and position~\cite{brincat2004underlying,hung2005fast,zoccolan2007trade,rust2010selectivity}). It was shown that different variations are differently treated trough the ventral visual pathway, for example responses to some variation (e.g., position) emerge before others (e.g., scale)~\cite{isik2014dynamics}. However, there is not any data addressing this for other variations. Yet depending on the type of variation, the visual system may use different sources of information to handle rapid object recognition. Therefore, the responses to each variation, separately, or in different combinations can provide valuable insight about how the visual system performs invariant object recognition. Because DCNNs claim to be bio-inspired, it is also relevant to check if their performance, when facing these transformations, correlates with that of humans.

Here, we performed several behavioral experiments to study the processing of different object variation dimensions through the visual system in terms of reaction time and categorization accuracy. To this end, we generated a series of image databases consisting of different object categories which varied in different combinations of four major variation dimensions: position, scale, in-depth and in-plane rotations. These databases were divided into three major groups:1) objects that varied in all four dimensions; 2) object that varied in combination of three dimensions (all possible combinations); 3) objects that varied only in a single dimension. In addition, each database has two background conditions: uniform gray and natural. Hence, our image database has several advantages for studying the invariant object recognition. First, it contains a large number of object images, changing across different types of variation such as geometric dimensions, object instance, and background. Second, we had a precise control over the amount of variations in each dimension which let us generate images with different degrees of complexity/difficulty. Therefore, it enabled us to scrutinize the behavior of humans, while the complexity of object variations gradually increases. Third, by eliminating dependencies between objects and backgrounds, we were able to study invariance, independent of contextual effects. 
 
Different combinations of object variations allowed us to investigate the role of each variation and combination in the task complexity and human performance. Interestingly, although different variations were linearly combined, the effects on reaction time and accuracy were not modulated in that way, suggesting that some dimensions substantially increased the task difficulty. The overall impression of our experimental results indicate that humans responded differently to different combination of variations, some  variations imposed more difficulty and required more processing time. Also, reaction times and categorization accuracies indicated that natural backgrounds significantly affects invariant object recognition. 

Results showed that 3D object rotation is the most difficult variation either in combination with others or by itself. In case of three-dimension experiments, subjects had high categorization accuracy when object were not rotated in-depth, while their accuracy significantly dropped in other three-dimension conditions. The situation was a similar for the reaction times: when the in-depth rotation was fixed across levels, the reaction time was shorter than the other conditions which objects were allowed to rotate in-depth. Although we expected that rotation in plane might be more difficult than scale, our results suggest the opposite. Possibly, changing the scale of the object might change the amount of information conveyed through the visual system which would affect the processing time and accuracy. Besides, the accuracy was very low when the objects were located on the center of the image but varied in other dimensions, while the accuracy was higher when we changed the object position and fixed any other dimensions. This suggests that subjects were  better able to tolerate variations in objects' position.

Moreover, we investigated whether these effects are related to low-level image features such as contrast and luminance. The results showed that the correlation between these features and reaction time and accuracy is very low and insignificant across all levels, type of variations, and objects. This suggests that although different variations affect the contrast and luminance, such low-level features have little effect on reaction time and accuracy.

We also performed ultra-rapid object categorization for the three-dimension databases with natural backgrounds, to see if our results depend on presentation condition or not. Moreover, to independently check the role of each individual dimension, we ran a one-dimension experiment in which objects were allowed to vary in only one dimension. These experiments  confirmed the results of our previous experiments. Besides, we evaluated two powerful DCNNs over three- and one-dimension databases which surprisingly achieved similar results to those of humans.
It suggests that humans have more difficulty for those variations which are computationally more difficult.

In addition to object transformations, background variation can also affect the categorization accuracy and time. Here, we observed that using natural images as object backgrounds seriously reduced the categorization accuracy and concurrently increased the reaction time. Importantly the backgrounds we used were quite irrelevant. We removed object-background dependency, to purely study the impacts of background on invariant object recognition. However, object-background dependency can be studied in future to investigate how contextual relevance between the target object and surrounding environment would affect the process of invariant object recognition (e.g.,~\cite{remy2013object,bar2004visual,harel2014task}).

Another limitation of our work, is that we did not assess the question of the extent to which previous experience is required for invariant recognition. Here, presumably both humans and DCNNs (through training) had extensive experience of the four classes we used (car, animal, ship, motorcycle) at different positions, scales, and with different viewing angles, and it is likely that this helped to develop the invariant responses. Importantly, studies have shown that difficult variations (rotation in depth) are solved in the brain later in development compared to easier ones~\cite{nishimura2014size}, suggesting that the brain needs more training to solve complex variations. It would be interesting to perform similar experiments as here with subjects of different ages, to unravel how invariance to different variations evolve through the development. 

During the last decades, models have attained some scale and position invariant. However, attempts for building a model invariant to 3D variations has been marginally successful. In particular, recently developed deep neural networks has shown merits in tolerating 2D and 3D variations~\cite{kheradpisheh2015deep,cadieu2014deep,ghodrati2014feedforward}. Certainly, comparing the responses of such models with humans (either behavioral or neural data) can give a better insight about their performance and structural characteristics. Hence, we did the same experiments on two of the best deep networks to see whether they treat different variations as humans do. It was previously shown that these networks can tolerate variations in similar order of the human feed-forward vision~\cite{kheradpisheh2015deep,cadieu2014deep}. Our results indicate that as humans they also have more difficulties with in-depth rotation and scale variation.

However, the human visual system extensively exploits feedback and recurrent information to refine and disambiguate the visual representation. Also, our vision is continues. Hence, the human visual system would have higher accuracies if it was allowed to use feedback information and continuous visual input. But deep networks lack such mechanism which could help them to increase their invariance and recognition ability. The future advances in deep networks should put more focus on feedback and continuous vision.

Finally, our results show that variation levels strongly modulate both humans’ and DCNNs’ recognition performances, especially for rotation in depth and scale. Therefore these variations should be controlled in all the image datasets used in vision research. Failure to do so may lead to noisy results, or even misleading ones. For example a category may appear easier to recognize than another one only because its variation levels happen to be small in a given dataset. We thus think that our methodology and image databases could be considered as benchmarks for investigating the power of any computational model in tolerating different object variations. Such results could then be compared with biological data (electrophysiology, fMRI, MEG, EEG) in terms of performance, but also representational dissimilarity~\cite{Kriegeskorte2008}. It would help computational modelers to systematically evaluate their models in fully controlled invariant object recognition tasks, and  could help them to improve the variation tolerance in their models, and to make them more human-like. 

\section*{Acknowledgments}
We would like to thank the High Performance Computing Center at Department of Computer Science of University of Tehran, for letting us perform our heavy and time-consuming calculations on their computing cluster.

\begin{footnotesize}

\bibliographystyle{elsarticle-num}
\bibliography{references}
\end{footnotesize}

\newpage
\onecolumn

\maketitle

\renewcommand\thefigure{S\arabic{figure}}
\setcounter{figure}{0}

\section*{Supplementary information}
\setcounter{section}{0}
\section{Two-category rapid presentation experiments}
\subsection{Psychophysics experiment}
In these experiments, subjects categorized rapidly presented images from two categories: car and animal. Each trial started with a fixation cross presented at the center of the screen for 500 {\it ms}. An image was then randomly selected from the image database and was presented for 25 {\it ms} (2 frames of a 80 Hz monitor) followed by a gray blank screen for 25 {\it ms} (ISI). Immediately after the blank screen, a $1/f$ noise mask image was presented for 100 {\it ms}. Subjects were asked to rapidly and accurately press one of the two keys, labeled on the computer keyboard, to indicate which object category was presented. The next trial was started immediately after the key press. There were two conditions which are explained as following:

\begin{itemize}
\item \textbf{All-dimension database:} Object images were selected from the all-dimension
database (see Image generation in the main manuscript). Subjects participated into two sessions: 1) Objects on a gray uniform background; 2) Objects on randomly selected natural backgrounds. In each session, subjects were presented with 320 object images (2 categories $\times$ 4 levels $\times$ 40 images from each category), divided into two blocks of 160 images. We collected the data of 17 sessions for each condition (i.e., uniform and natural backgrounds).  

\item \textbf{Three-dimension database:} In this experiment, we used the three-dimension databases.This allowed us to study the effect of excluding the variations across one dimension on human performance in invariant object categorization: if the fixed dimension is more difficult than the others, subjects will be able to categorize the objects more accurately and within shorter time than if the fixed dimension is easier. In each session, subjects were presented with 960 images: 2 categories $\times$ 4 levels $\times$ 4 conditions ($\Delta_{Sc} = 0$, $\Delta_{Po} = 0$, $\Delta_{RP} = 0$, and $\Delta_{RD} = 0$) $\times$ 30 images per category. Note that we inter-mixed images of all conditions; so, subjects were unaware of the type of variations. Images were presented in four consecutive blocks of 240 images. We recorded 17 sessions for each background condition (i.e., objects on uniform and natural backgrounds). 

\end{itemize}

\subsection{Behavioral results}
\begin{figure}[!t]
\centering
\includegraphics[scale=1.8]{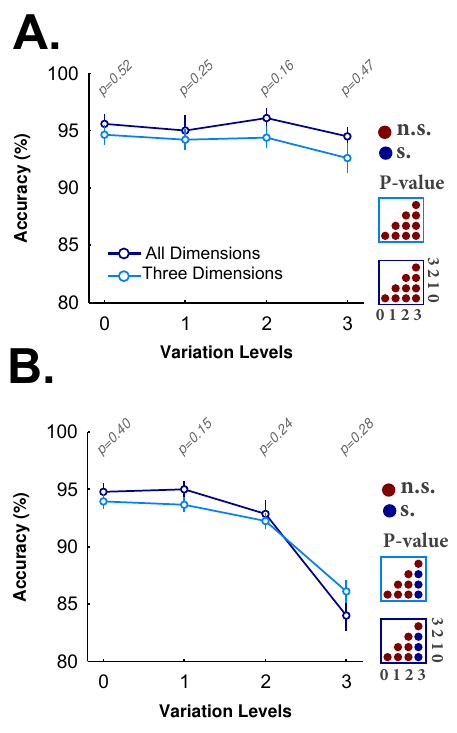}
\caption{\textbf{Human accuracy in two-category rapid invariant object categorization task.} A. The average accuracy of human subjects in categorization of car versus animal images, when objects had uniform background. The dark, blue curve shows the accuracy when objects varied across all dimensions and the light, blue curve demonstrates the accuracy when objects varied across three dimensions. Error bars are the standard deviation (STD). P values, printed at the top of curves, show whether the accuracy between all- and three-dimension experiments significantly differ (Wilcoxon rank sum test). Color-coded matrices, at the right, show all possible pair-wise comparisons across levels, indicating  whether changes in accuracy were statistically significant (Wilcoxon rank sum test; each matrix corresponds to one curve; see color of the frame). B. Categorization accuracies when objects had natural backgrounds.}
\label{figure_S1}
\end{figure}

Subjects achieved remarkably high accuracy in  categorization of rapidly presented object images from two categories while they varied across different dimensions (car versus animal; see the experimental settings in previous Section). When objects had uniform background, the average accuracy of subjects across different variation levels was about $~95\%$ (Figure~\ref{figure_S1}.A). There was no significant difference between the accuracies when  objects varied across all and three dimensions (Wilcoxon rank-sum test). Also, there was no significant accuracy drop across the variation levels in both experimental conditions (see the color-coded p-value matrices at the right side of Figure~\ref{figure_S1}.A, Wilcoxon rank sum test).

However, in the experiment with natural backgrounds, there was a significant accuracy drop as the variation level increased in both all- and three-dimension conditions (see the color-coded p-value matrices in the right side of Figure~\ref{figure_S1}.B). This shows that the presence of distractors in the background dramatically affects the accuracy of subjects in invariant object recognition, specifically at higher variation levels. 

\begin{figure*}[!htb]
\centering
\includegraphics[scale=.9]{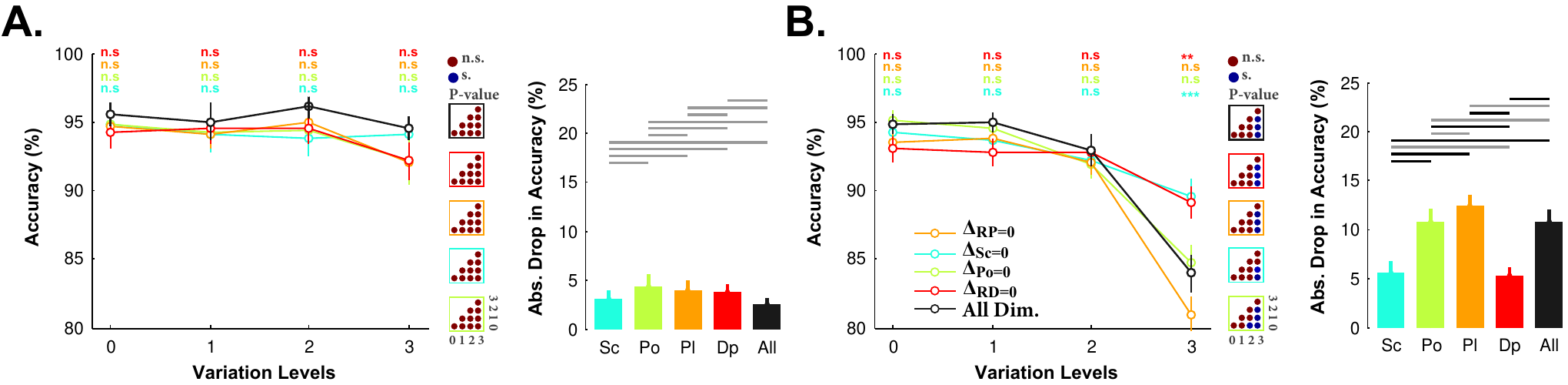}
\caption{\textbf{Human accuracy in two-category rapid invariant object categorization task in different three-dimension conditions.} A. Left, human average accuracy in categorization of car versus animal images, when objects had uniform backgrounds. Each curve corresponds to one condition: $\Delta_{Sc}=0 $, $ \Delta_{Po}=0 $, $ \Delta_{RP}=0 $, $ \Delta_{RD}=0 $ (as specified with different colors). Error bars are the standard deviation (STD). P values, depicted on the top of curves, show whether the accuracy between all-dimension and three-dimension conditions significantly differ (Wilcoxon rank sum test). Color-coded matrices show whether changes in accuracy across levels are statistically significant (Wilcoxon rank sum test; each matrix corresponds to one curve; see color of the frame). Right, absolute accuracy drop between level 0 and level 3 (mean+/-STD). Each bar corresponds to one condition. The horizontal lines on the top of bar plot show whether the differences are significant (gray line: insignificant, black line: significant). B. Categorization accuracy when objects had natural backgrounds.}
\label{figure_S2}
\end{figure*}

Figure~\ref{figure_S1} showed the overall accuracy in two experimental conditions, but it did not show how accuracy depends on variations across different dimensions. For this purpose, we computed the average accuracies for different conditions in the three-dimension experiment (i.e., $ \Delta_{Sc}=0 $, $ \Delta_{Po}=0 $, $ \Delta_{RP}=0 $, $ \Delta_{RD}=0 $) and compared them with the all-dimension case. Figure~\ref{figure_S2}.A shows that when objects had uniform background, there was no significant difference in the accuracies of different three-dimension conditions, suggesting that subjects could robustly categorize objects in this case, even at high variation levels. The accuracy drop between level 0 and level 3 was also very small (see the bar plot in Figure~\ref{figure_S2}.A). 

The situation was completely different when objects had natural backgrounds. Figure~\ref{figure_S2}.B illustrates that although there is no significant difference in accuracies between three-dimension conditions at low and intermediate variation levels, there is a significant difference  (almost 15\%) between  the accuracies in $ \Delta_{RD}=0 $ and $ \Delta_{Sc}=0 $, and $ \Delta_{RP}=0 $ and $ \Delta_{Po}=0 $ at the highest variation level (level 3). This is also evident in the absolute accuracy drops, which indicates that the accuracy drop in $ \Delta_{RD}=0 $ and $ \Delta_{Sc}=0 $ was significantly smaller than $ \Delta_{RP}=0 $ and $ \Delta_{Po}=0 $ (Figure~\ref{figure_S2}.B, bar plot). These suggest that the presence of in-depth rotation and scale variation made the object recognition very difficult. The accuracy in all-dimension experiment was similar to $ \Delta_{RP}=0 $ and $ \Delta_{Po}=0 $ conditions (See p values at the top of Figure~\ref{figure_S2}). 

\begin{figure}[!b]
\centering
\includegraphics[scale=2]{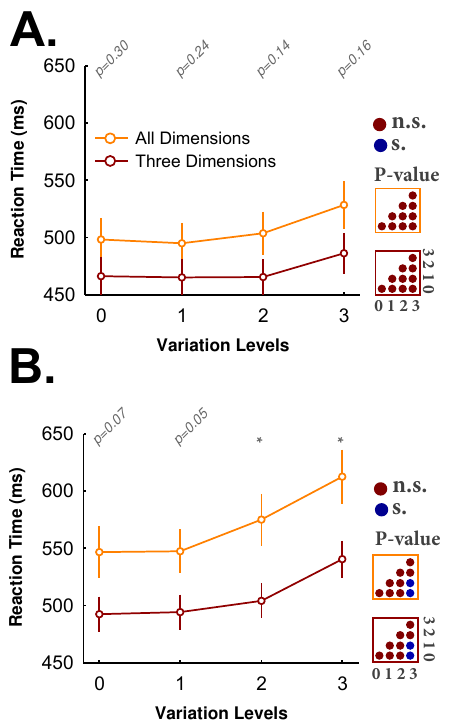}
\caption{\textbf{Average reaction time of humans in two-category rapid invariant object categorization task.}  A. Average and standard error of the mean (SEM) of subjects' reaction time in all- and three-dimension conditions, when objects had uniform background.  The orange curve (resp. brown) shows the reaction time when objects varied in all (resp. three) dimensions. Note that reaction times in three-dimension case are the overall reaction times across different conditions. P values, depicted at the top of curves, show whether the reaction time difference between all- and three-dimension are significant (Wilcoxon rank sum test). Color-coded matrices, at the right, show all possible pair-wise comparisons across levels, indicating that whether or not reaction time changes are statistically significant (Wilcoxon rank sum test; each matrix corresponds to one curve; see color of the frame) B. Reaction times when objects had natural background. }
\label{figure_S3}
\end{figure}

We also recorded the reaction times of subjects performing two-category experiments. Here, we first report the overall reaction time of human subjects regardless of the type of variations.  Figures~\ref{figure_S3}.A and~\ref{figure_S3}.B present the reaction times of different variation levels when objects varied in all and three dimensions for uniform and natural backgrounds, respectively.

The average reaction time in the all-dimension condition is longer than in the three-dimension condition across all levels, although the differences are not significant (Figure~\ref{figure_S3}.A, wilcoxon rank sum). In addition, for both all-  and three-dimension condition, there is no significant change in reaction times across the variation levels (see the color-coded p-value matrices in the right side of Figure~\ref{figure_S3}.A; they show all possible pair-wise comparisons across levels; wilcoxon rank sum test).  As shown in Figure~\ref{figure_S1}.A, results for the accuracies  are similar. Hence, it can be said that humans can accurately ($~95\%$) and quickly ($~450-520$ ms) categorize varied object images in our two-category invariant object categorization tasks with uniform background.

The general reaction times for natural background condition are provided in Figure~\ref{figure_S3}.B. As can be seen, there are significant increments in reaction times of both all- and three-dimension, specifically in higher variation levels. The trend for the accuracies is similar(see Figure~\ref{figure_S1}.B): the accuracies significantly drop in higher variation levels.  These together show that distractors in clutter backgrounds significantly affect the performance  of humans in recognition of highly varied objects.  Moreover, although there was no significant difference in the accuracies between all- and three-dimension experiments, the reaction times were significantly different at variation levels 2 and 3, indicating that objects with variations across all-dimensions needed more processing time than in the three-dimension case.

\begin{figure*}[!t]
\centering
\includegraphics[scale=.9]{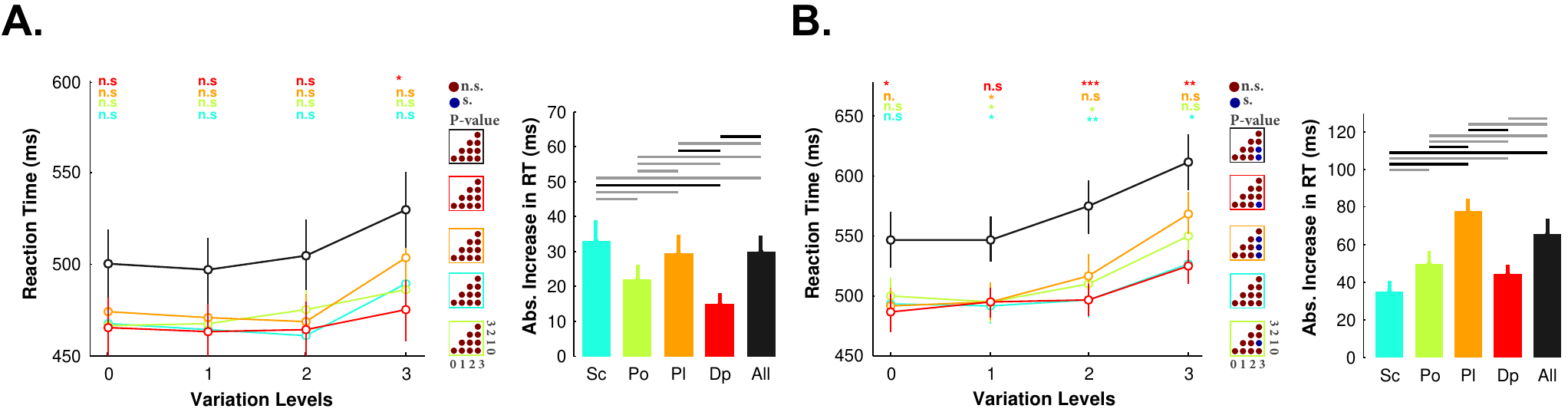}
\caption{\textbf{Average reaction time of humans in two-category rapid invariant object categorization task for different three-dimension conditions.} A. Left, average and standard error of the mean (SEM) of subjects' reaction time in all-dimension and different three-dimension conditions, when objects had uniform background.  Each curve corresponds to one condition: $\Delta_{Sc}=0 $, $ \Delta_{Po}=0 $, $ \Delta_{RP}=0 $, $ \Delta_{RD}=0 $ (as specified with different colors).  P values, depicted on the top show whether the reaction time difference between all-dimension and other three-dimension conditions are significant (Wilcoxon rank sum test). Color-coded matrices show all possible pair-wise comparisons across levels, indicating whether or not the reaction time changes in each condition are statistically significant (Wilcoxon rank sum test; each matrix corresponds to one curve; see color of the frame). Right, absolute reaction time increase between level 0 and level 3 (mean+/-STD). The horizontal lines on the top show whether the differences are significant (gray line: insignificant, black line: significant). B. Average and SEM of subjects' reaction time in all- and different three-dimension conditions, when objects had natural backgrounds.}
\label{figure_S4}
\end{figure*}

Figures~\ref{figure_S4}.A and~\ref{figure_S4}.B  demonstrate the reaction times of each variation combination in the three-dimension case as well as in the all-dimension case across different levels for both uniform and natural backgrounds, respectively. Comparing the reaction times in different three-dimension conditions with uniform background shows insignificant difference among them (Figure~\ref{figure_S4}.A). It means that the elimination of any dimension dose not affect much the reaction time. In other words, it is easy for humans to categorize rapidly presented car and animal images with uniform gray background even in high variation levels independently of the type of variations. However, we can see, from the left bar plot, that removing rotation in-depth (red bar) made the task easier with smaller absolute drop in reaction time from level 0 to level 3 (this is significant comparing to other conditions, see color-coded horizontal lines at the top of the bar plots). It is also evident that the all-dimension case has a higher reaction time and it is the most difficult task.

However, as shown in Figure~\ref{figure_S4}.B, the situation is different when objects had natural backgrounds. The reaction times of $ \Delta_{RP}=0 $ and $ \Delta_{Po}=0 $ are longer than  $ \Delta_{RD}=0 $ and $ \Delta_{Sc}=0 $. As seen in Figure~\ref{figure_S2}.B, the accuracies of $ \Delta_{RP}=0 $ and $ \Delta_{Po}=0 $ were also significantly lower than that of the other two three-dimension conditions. This confirms that in-depth rotation and scale variations are more difficult than the other two and need more processing time (Figure~\ref{figure_S4}.B). It is also noteworthy that the reaction times in the all-dimension condition are significantly longer than in all the three-dimension conditions (See p values at the top of Figure~\ref{figure_S4}.B). 

\section{Four-category rapid presentation experiments}
\subsection{Category-wise accuracy}
\begin{figure*}[!htb]
 \centering
 \includegraphics[scale=0.85]{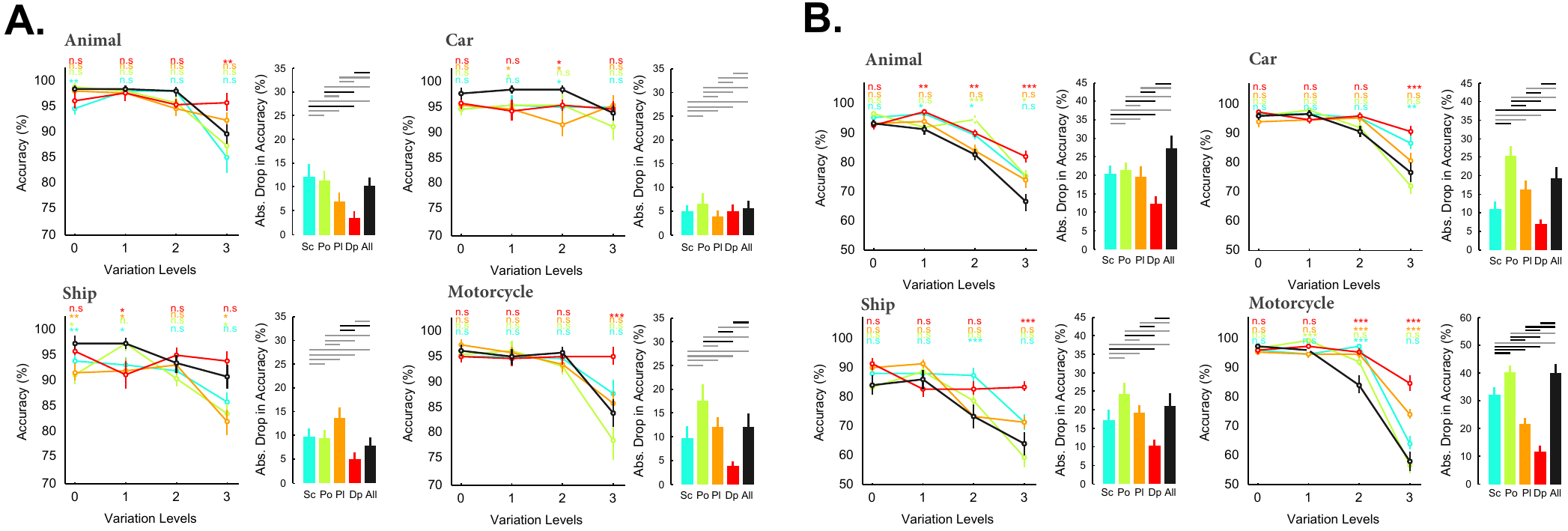}
 \caption{\textbf{Categorization accuracy and absolute drop in accuracy for different conditions and object categories.} A. Uniform background. The left plot for each object category illustrates the accuracies of different three-dimension conditions. Error bars are the standard deviation (STD). P values, depicted on the top show whether the accuracy between all-dimension and other three-dimension conditions are significantly different (Wilcoxon rank sum test). For each object category, the bar plot on the right demonstrates the absolute accuracy drop between level 0 and level 3 (mean+/-STD). The horizontal lines on the top of these bar plots show whether the differences between variation conditions are significant (gray line: insignificant, black line: significant). B. Natural background, the conventions are identical to A.}
 \label{figure_S5}
 \end{figure*}

The global accuracies of humans for the four-category rapid experiments are presented in the main manuscript. Here, for each category, we present the human accuracies in different variation and background conditions. Figure~\ref{figure_S5}.A illustrates the category-wise accuracies in case of uniform background. As seen, the $ \Delta_{RD}=0 $ condition has the highest accuracy in almost all categories, even at the highest variation level, while $ \Delta_{Po}=0 $ has the lowest accuracy. These two conditions also have the lowest and highest accuracy drops, respectively. Generally, subjects made the maximum error while categorizing motorcycle and ship categories, specifically at the highest variation level, while they achieved the greatest accuracy in categorization of car images. This is also reflected in the accuracy drop bar plots with the highest drop in categorization of motorcycle images and the lowest drop for car images. This means that subjects could better tolerate variations in car images and they have more difficulty to deal with variations in motorcycle and ship instances. 

Figure~\ref{figure_S5}.B represents the accuracies of each category in a separate plot, for the natural background experiment. Although the trend in accuracy is different across categories, it is evident that $\Delta_{Sc}=0 $ and $\Delta_{RD}=0$ were the easiest in three-dimension on conditions. Moreover, subjects had higher errors (and longer reaction times; see Figure~\ref{figure_S11}.B) in categorization of ships and motorcycles compared to the other categories. This can be seen in the bar plots of Figure~\ref{figure_S5}.B, where the highest accuracy drop was observed when categorizing images from the motorcycle category. In contrast, we observed the lowest accuracy drop when categorizing car images. This indicates that, contrary to motorcycle and ship categories, subjects could better tolerate variations in cars. The most difficult categorization task was when position variation was set to 0 (green curves and bars). In this case, objects varied in the dimensions (i.e., scale, in-depth and in-plane rotations) that imposed more difficulty to the task.

\subsection{Confusion matrix analysis}
\begin{figure*}[!t]
\centering
\includegraphics[scale=1.3]{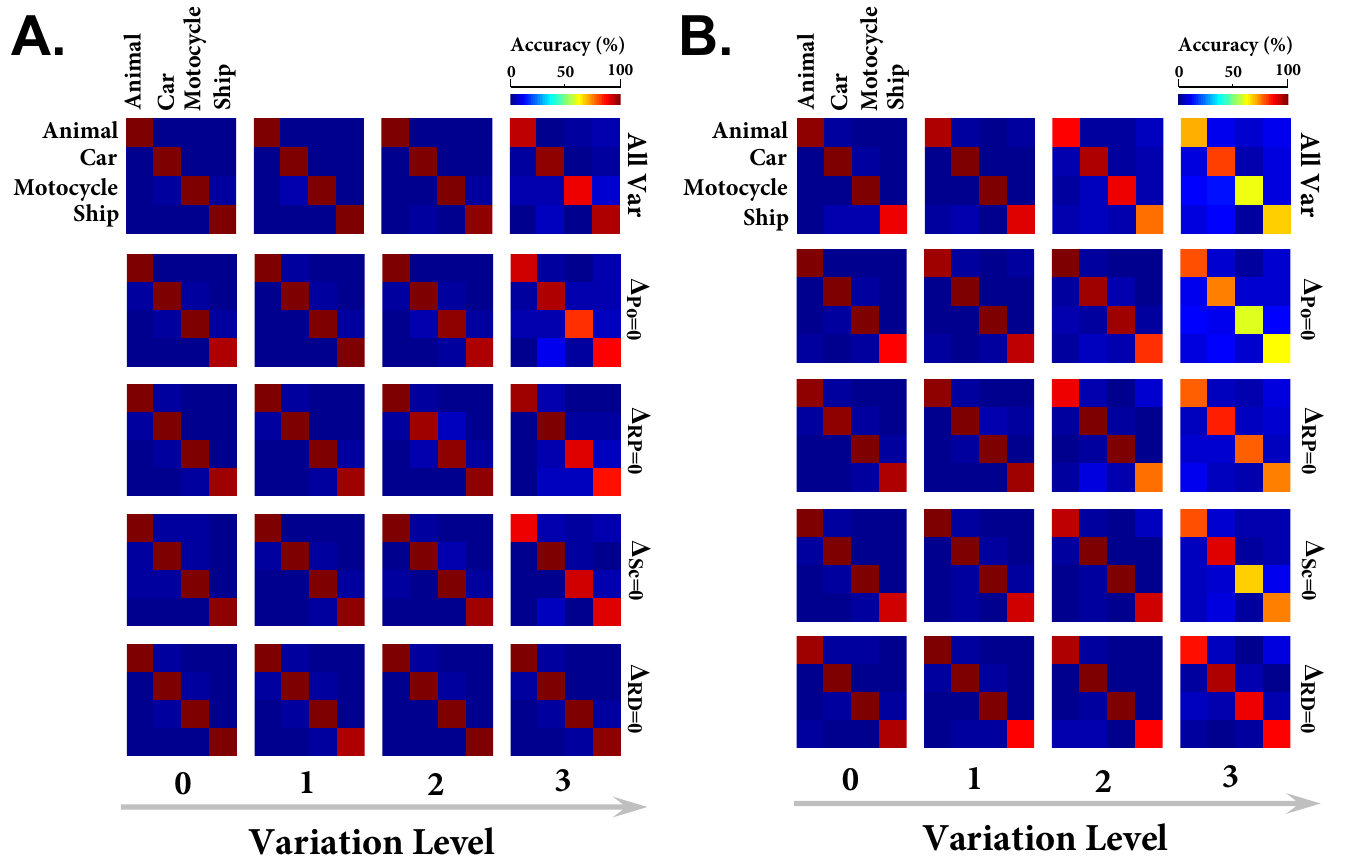}
\caption{\textbf{Confusion matrices for rapid object categorization tasks for all- and three-dimension conditions.} A. Confusion matrices when objects had uniform backgrounds. Each column of confusion matrices corresponds to a variation level and each row refers to an experimental condition (written at the right end). The name of categories is written at the first, top-left confusion matrix. The color bar at the top-right indicates the range of accuracies. B. Confusion matrices when object has natural backgrounds.}
\label{figure_S6}
\end{figure*}

To have a closer look at error rate and miscategorization across categories, we calculated the confusion matrices of all- and three-dimension experiments. Figure~\ref{figure_S6}.A shows that in the uniform background condition, the categorization error increased at level 3, with the highest error rate when categorizing ship and motorcycle images (e.g., most of wrongly assigned ship labels corresponded motorcycles images). Comparing different three-dimension conditions shows that the miscategorization rate in $\Delta_{Po}=0 $ was higher than in the other conditions while the lowest miscategorization rate was observed in $\Delta_{RD}=0$. The miscategorization rate in natural backgrounds experiment was higher than in the uniform background condition (Figure ~\ref{figure_S6}.B), even at low and intermediate variation levels. The error rate was lower in $\Delta_{RD}=0$ than in the other conditions. On the other hand, the highest error rate was observed in $\Delta_{Po}=0 $.

\subsection{Contrast and luminance analysis}
As a control, we computed the correlation between low-level image statistics (contrast and luminance) and the performance of human subjects. Here, we investigate whether the changing pattern of human accuracy is due to the nature of the task or to the changes in image statistics. The image contrast is computed by the Root Mean Square Contrast method which is defined as the standard deviation of the intensity values of all pixels in the image divided by the mean intensity, and the image mean luminance is obtained by averaging the pixel intensities.

\begin{figure}[!htb]
\centering
\includegraphics[scale=.5]{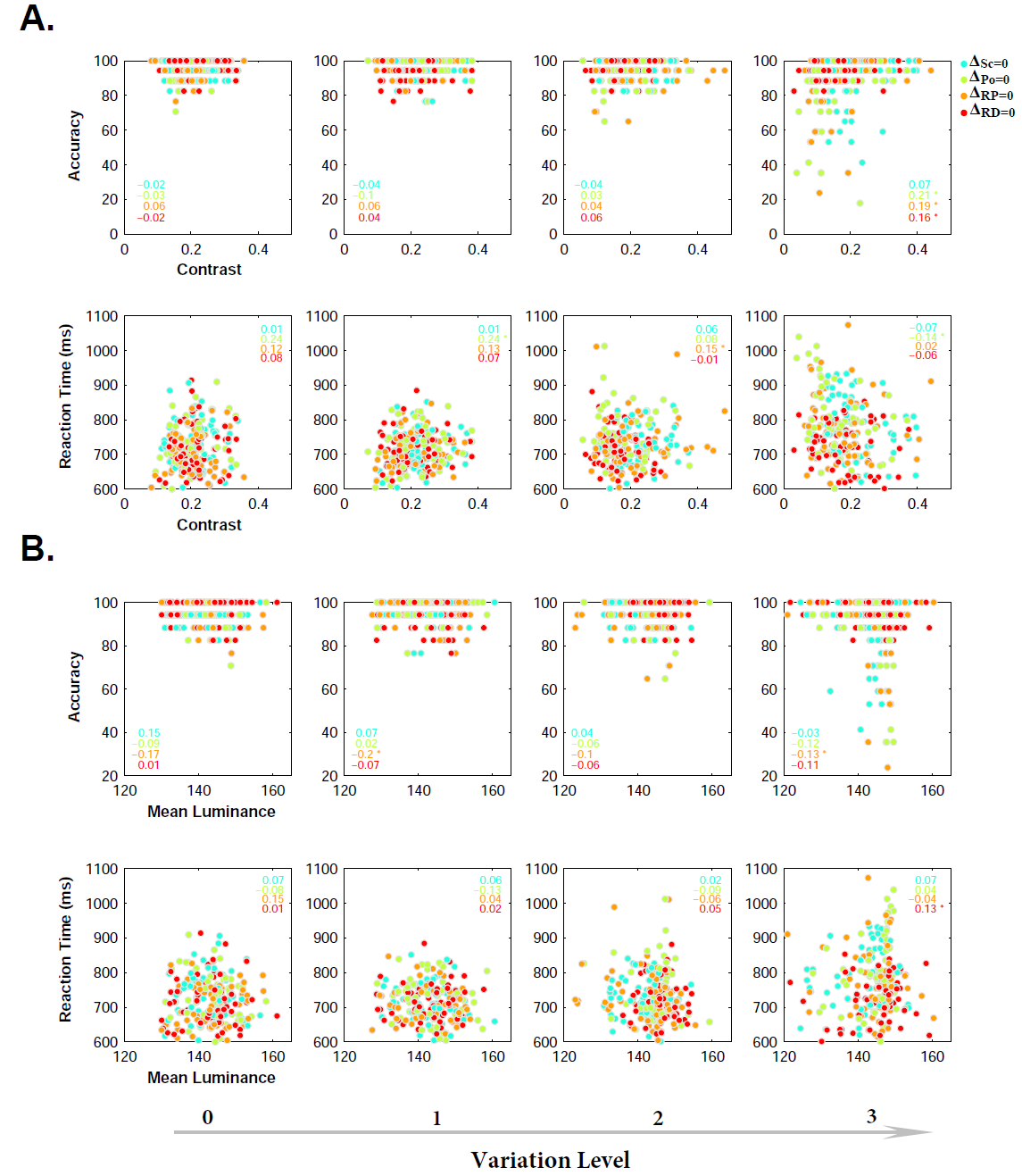}
\caption{\textbf{The correlation between low-level statistics (contrast and luminance) and the performance of human subjects, when objects had uniform backgrounds.} A. Correlation between the contrast of images (root mean square contrast) and human accuracy (top row) and reaction time (bottom row) for all levels and three variation conditions. Each point refers to an image and colors indicate an experimental condition. Correlation values are depicted in each scatter plot with corresponding colors (Pearson correlation). Significant correlations are specified using asterisks next to numbers. Scatter plots are plotted for all levels from level 0 (left) to level 1 (right). B.  Correlation between the luminance of images (mean luminance of all pixels) and human accuracy (top row) and reaction time (bottom row) for all levels and three variation conditions.}
\label{figure_S7}
\end{figure}

Figures~\ref{figure_S7}.A and~\ref{figure_S7}.B, respectively, illustrate the correlation values of image contrast and mean luminance with human accuracy and reaction time over different variation levels and variation conditions for the three-dimension uniform background task. As can be seen, the correlation of image contrast with human accuracy and reaction time are negligible and insignificant for almost all variation conditions. The situation is similar for image mean luminance. These together indicate that, for the uniform background task, image statistics  such as contrast and luminance do not significantly contribute to humans' invariant object recognition process.

\begin{figure}[!htb]
\centering
\includegraphics[scale=.5]{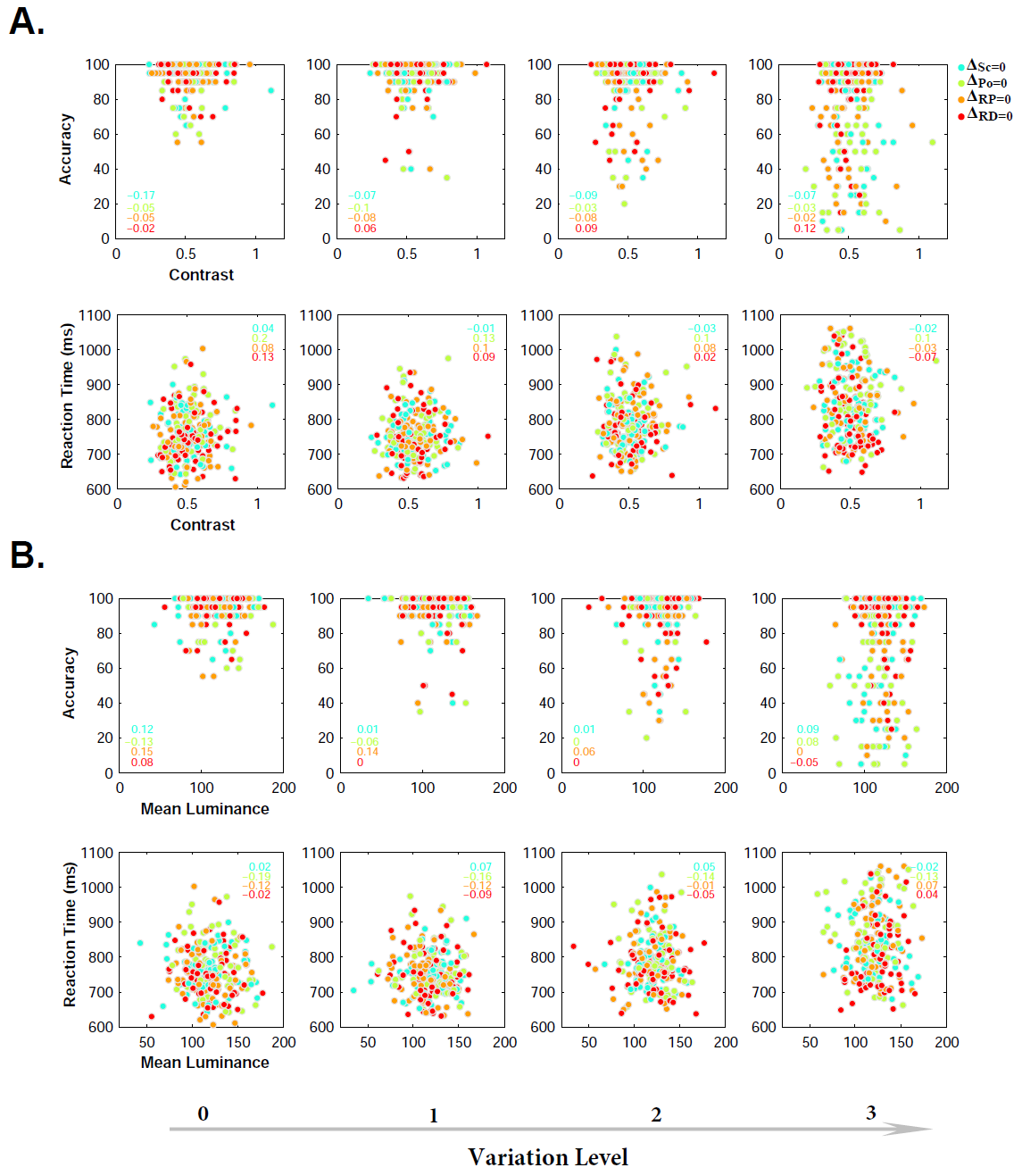}
\caption{\textbf{The correlation between low-level statistics (contrast and luminance) and the performance of human subjects, when objects had natural backgrounds.} A. Correlation between the contrast of images (root mean square contrast) and human accuracy (top row) and reaction time (bottom row) for all levels and three variation conditions. Each point refers to an image and colors indicate an experimental condition. Correlation values are depicted in each scatter plot with corresponding colors (Pearson correlation). Significant correlations are specified using asterisks next to numbers. Scatter plots are plotted for all levels from level 0 (left) to level 1 (right). B.  Correlation between the luminance of images (mean luminance of all pixels) and human accuracy (top row) and reaction time (bottom row) for all levels and three variation conditions.}
\label{figure_S8}
\end{figure}

We did the same analysis for the three-dimension natural background conditions. The correlation values of image contrast and mean luminance with human accuracy and reaction time over different variation levels and variation conditions are presented in Figure~\ref{figure_S8}.A and Figure~\ref{figure_S8}.B, respectively. As shown in these figures, the correlation values for both contrast and mean luminance are small and statistically insignificant for all variation levels and variation conditions. This means that changes in luminance or contrast do not affect human accuracy and reaction time in invariant object recognition.

\subsection{Reaction time}
\begin{figure}[!t]
\centering
\includegraphics[scale=2]{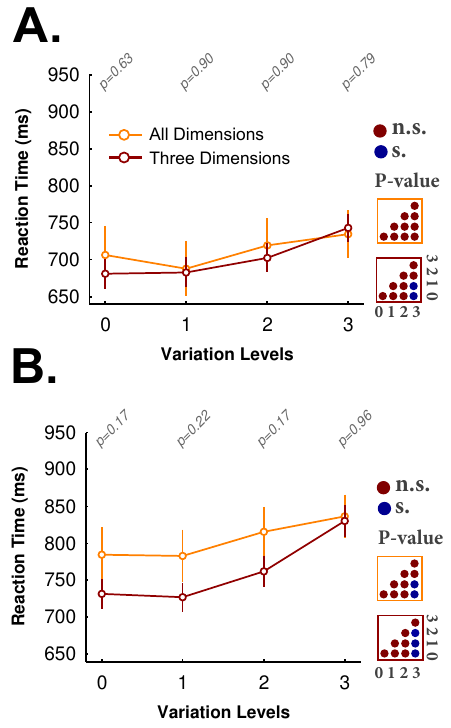}
\caption{\textbf{Average reaction time of human subjects in rapid invariant object categorization task.}  A. Average and standard error of the mean (SEM) of subjects' reaction time in the all- and three-dimension conditions, when objects had uniform background.  The orange curve shows the reaction time when objects varied in all dimensions and the brown curve demonstrates the data for three dimensions.  P values, depicted on the top, show whether the reaction time difference between the all- and three-dimension experiments are significant (Wilcoxon rank sum test). Color-coded matrices, on the right, show all possible pair-wise comparisons across levels, indicating whether or not the reaction time changes are statistically significant (Wilcoxon rank sum test; each matrix corresponds to one curve; see color of the frame). B. Average and SEM of subjects' reaction time in the all- and three-dimension conditions, when objects had natural backgrounds.}
\label{figure_S9}
\end{figure}

Figures~\ref{figure_S9}.A and~\ref{figure_S9}.B demonstrate the average reaction time of human subjects over different variation levels of all- and three-dimension experiments, for uniform and natural backgrounds, respectively. Note that in these figures the reaction time of three-dimension conditions are averaged over different variation types, meaning that we did not break into the type of variation.

As seen in Figure~\ref{figure_S9}.A, in case of uniform background the reaction times of the all- and three-dimension experiments are very close to each other and there is no significant difference between these conditions. However, humans needs more processing time ($\sim 50$ms) for the same tasks but with natural backgrounds (see Figure~\ref{figure_S9}.B). Contrary to the uniform background case, there is  a big difference (although not statistically significant) between the average reaction times of the all- and three-dimension experiments with natural backgrounds.

\begin{figure*}[!htb]
\centering
\includegraphics[scale=1.4]{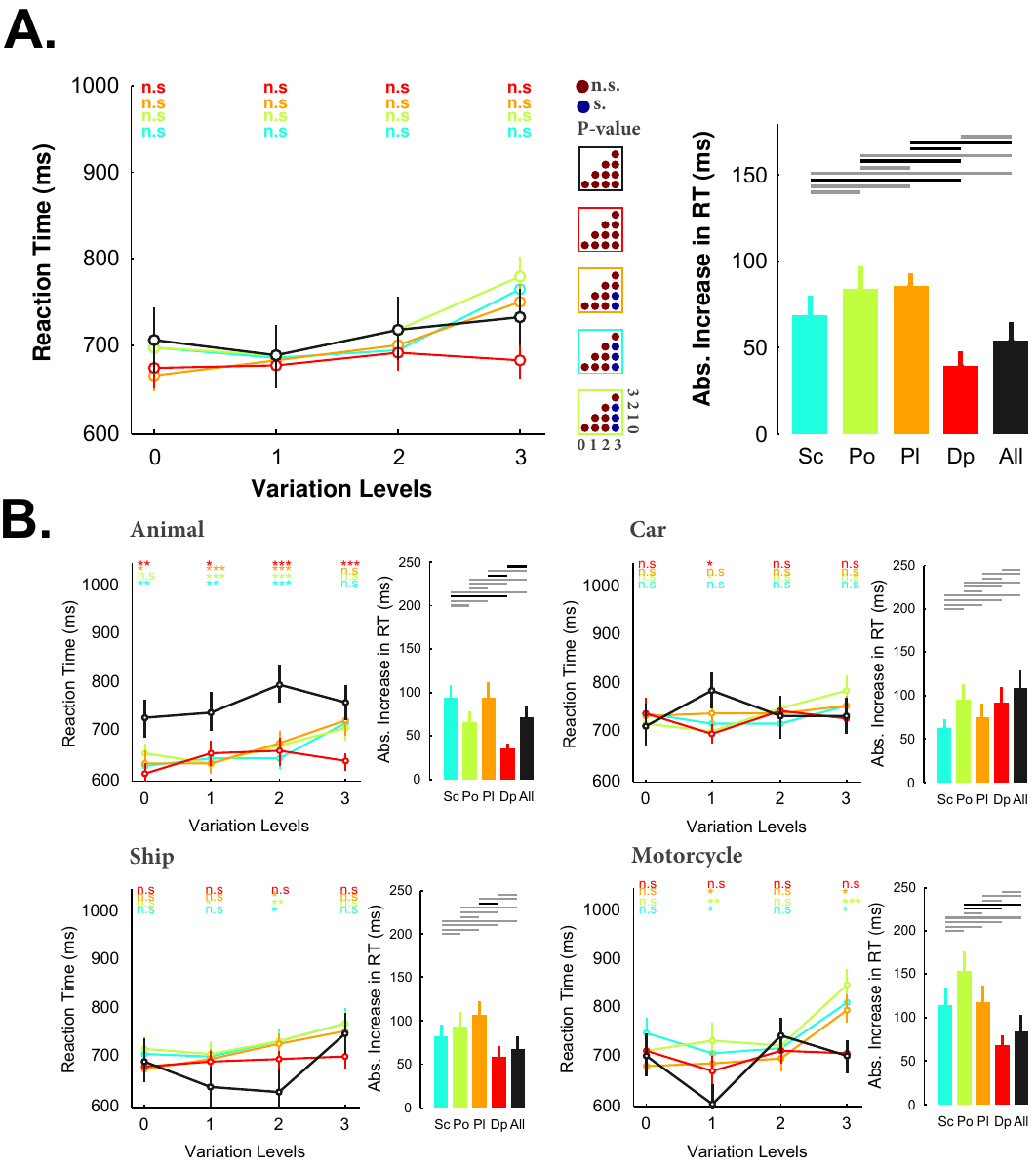}
\caption{\textbf{Average reaction time of human subjects in rapid invariant object categorization task for the all-dimension and different three-dimension conditions, when objects had uniform backgrounds.} A. Left, average and standard error of the mean of subjects' reaction time in the all-dimension and different three-dimension conditions, when objects had uniform backgrounds.  Each color refers to a condition. P values, depicted on the top, show whether the reaction time difference between the all-dimension and the  other three-dimension conditions are statistically significant (Wilcoxon rank sum test). Color-coded matrices, on the right, show all possible pair-wise comparisons across levels, indicating whether the reaction time changes in each condition are statistically significant (Wilcoxon rank sum test; each matrix corresponds to one curve; see color of the frame). Right, absolute increase in reaction time between level 0 and level 3 (mean+/-STD). The horizontal lines on the top show whether the differences are significant (gray line: insignificant, black line: significant). B. Reaction time and absolute increase in reaction time for different conditions and object categories.}
\label{figure_S10}
\end{figure*}

Figure~\ref{figure_S10}.A provides the reaction times of each three-dimension condition as well as of the all-dimension experiment, for the case of uniform background. In low and intermediate variation levels, the reaction times of the different three-dimension conditions as well as of the all-dimension one are closed to each other. However, in the most difficult level, the reaction times of $ \Delta_{Sc}=0 $, $ \Delta_{Po}=0 $, and $ \Delta_{RP}=0 $ significantly grows up, while it does not change in $ \Delta_{RD}=0 $ (see color-coded matrices). Indeed, when an object is not rotated in-depth humans can more quickly categorize it than when another dimension is fixed and the object is allowed to rotate in-depth. This again indicates that humans need more processing time to categorize depth-rotated objects. Evidently, $ \Delta_{Po}=0 $ has higher reaction times than the other conditions, specifically in the highest level. In other words, if we do not change the position of the object but vary it in other dimensions, subjects need more time to categorize it. This means that position variation is easier to overcome for humans. Also, these results are confirmed by comparing the absolute increase in reaction times of different conditions presented in the bar plot of Figure~\ref{figure_S10}.A.

\begin{figure*}[!htb]
\centering
\includegraphics[scale=1.4]{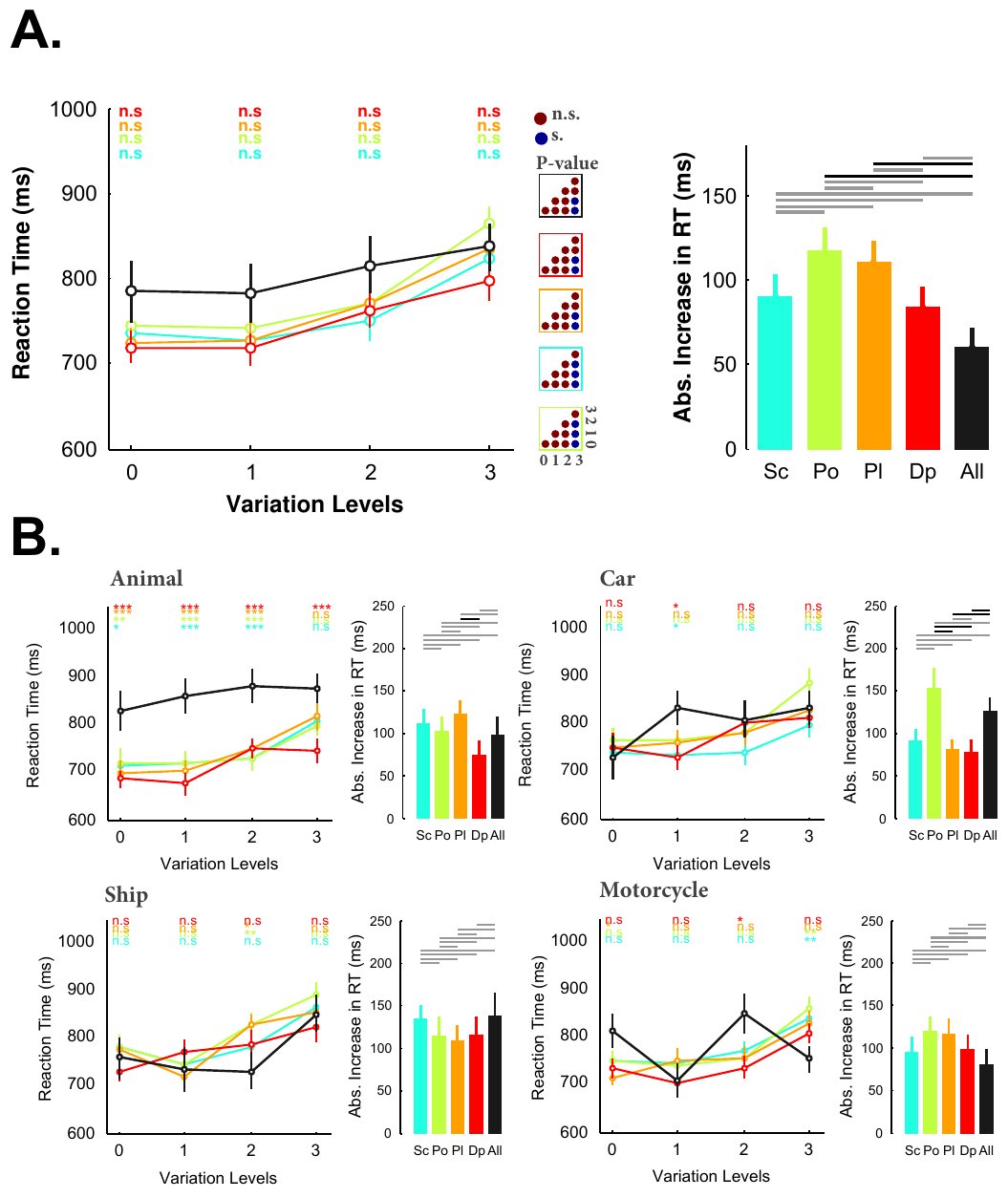}
\caption{\textbf{Average reaction time of human subjects in rapid invariant object categorization task for the all-dimension and different three-dimension conditions, when objects had natural backgrounds.}  A. Left, average and standard error of the mean of subjects' reaction time in the all-dimension and different three-dimension conditions, when objects had natural backgrounds.  Each color refers to a three-dimension condition (p values and matrices were calculated using a similar approach to fig~\ref{figure_S10}). Right, absolute increase in reaction time between level 0 and level 3 (mean+/-STD). The horizontal lines on the top show whether the differences are significant (gray line: insignificant, black line: significant). B.  Reaction time and absolute increase in reaction time for different conditions and object categories.}
\label{figure_S11}
\end{figure*}

For each object category, the reaction times of the all- and three-dimension uniform background experiments  are presented in a separate plot in Figure~\ref{figure_S10}.B. In line  with categorization accuracies (see Figure~\ref{figure_S5}.A),  $ \Delta_{RD}=0 $ has the shortest and $ \Delta_{Po}=0 $ has the longest reaction times in almost all categories. By the way, Motorcycles have the highest reaction times specifically in level 3, and also have the greatest increase in reaction time. Figure ~\ref{figure_S11}.A illustrates the reaction time and absolute increase in reaction time of the all- and three-dimension experiments with natural backgrounds. The category-wise reaction times and corresponding reaction time increments from the lowest to the highest variation level are shown in Figure~\ref{figure_S11}.B.

\section{Four-category ultra-rapid presentation experiments (reaction time)}
\begin{figure*}[!htb]
\centering
\includegraphics[scale=1.2]{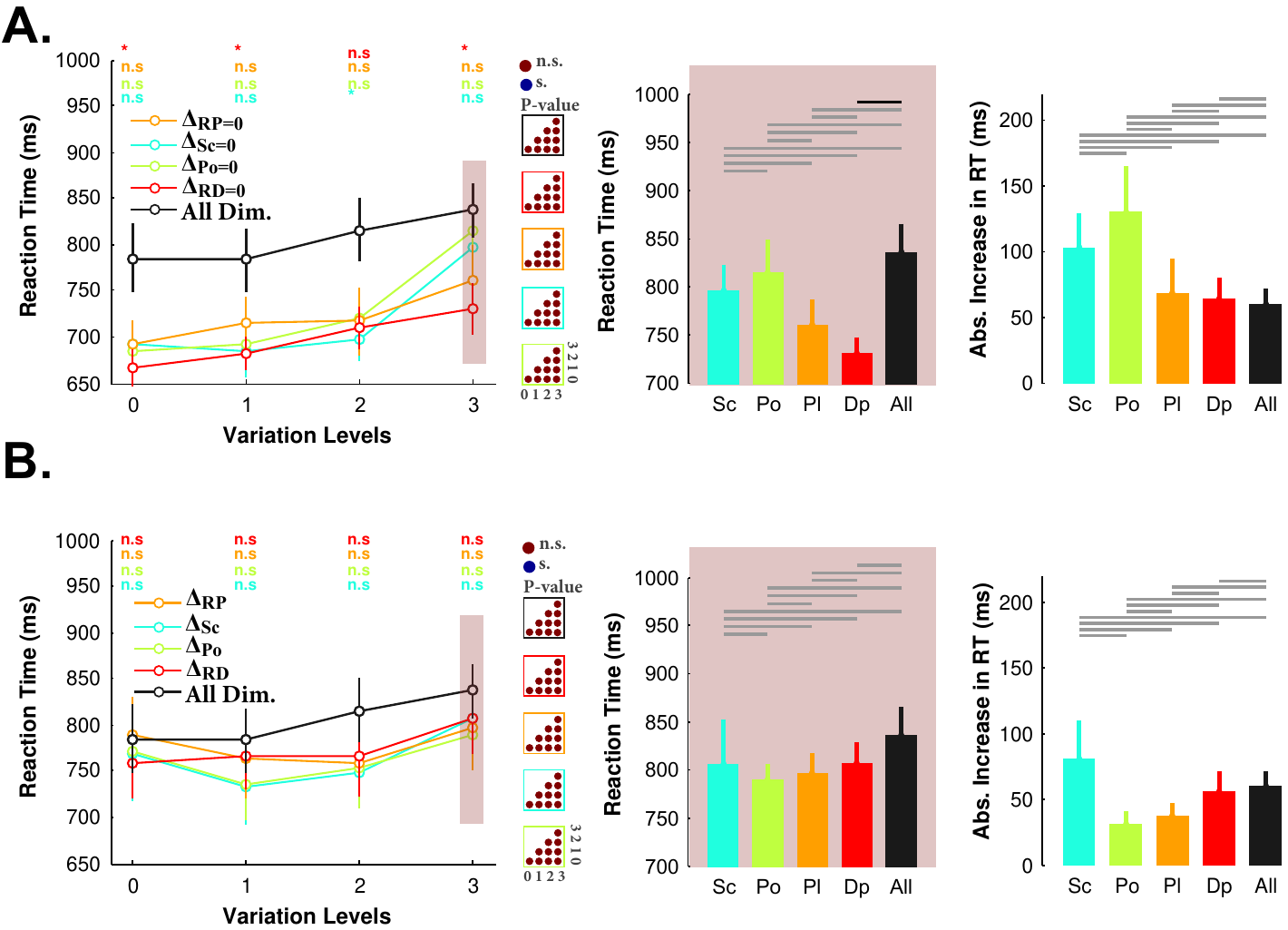}
\caption{\textbf{Reaction time of human subjects in ultra-rapid invariant object categorization task for the different three-dimension and one-dimension conditions, when objects had natural backgrounds.}  A. Left, average and standard error of the mean of subjects' reaction time in different three-dimension conditions. Each curve corresponds to one condition: $\Delta_{Sc}=0 $, $ \Delta_{Po}=0 $, $ \Delta_{RP}=0 $, $ \Delta_{RD}=0 $ (as specified with different colors). Horizontal axis shows variation levels from level 0-3.  Error bars are the standard deviation (STD). Color-coded matrices, on the right, show all possible pair-wise comparisons across levels, indicating whether the reaction time changes in each condition are statistically significant (Wilcoxon rank sum test; each matrix corresponds to one curve; see color of the frame). Middle, reaction times at the most difficult level for different three variation conditions (each bar corresponds to one condition). The horizontal lines on the top shows whether the differences are significant (gray line: insignificant, black line: significant). Right, absolute increase in reaction time between level 0 and level 3 (mean+/-STD). B. Left, average and standard error of the mean of subjects' reaction time in different one-dimension conditions (details of the plot are similar to A).  Middle, reaction times at the most difficult level for different three variation conditions (each bar corresponds to a condition). Right, absolute increase in reaction time between level 0 and level 3 (mean+/-STD). }
\label{figure_S12}
\end{figure*}

The accuracies of humans for these experiments are available in the main manuscript, and here we present the  reaction times. The left plot in Figure~\ref{figure_S12}.A illustrates the reaction times of the ultra-rapid  invariant object categorization task for the three-dimension conditions, when objects had natural backgrounds. The absolute reaction time increase, from the first to the last variation level, as well as the reaction time in level 3 are also presented in the middle and the right plot of Figure~\ref{figure_S12}.B, respectively. Although there is no statistically significant difference in the reaction times of different conditions, $\Delta_{RD}=0$ has the lowest average reaction time in level 3 (note that these results are the average of five subjects only, so, small sample size might be the reason for insignificant differences). Also, $\Delta_{RD}=0$  has the smallest increase in reaction time from the first the highest level. Once again, $\Delta_{Po}=0$ has the highest average reaction time (although insignificant) in level 3, and the largest increase in reaction time from level 0 to level 3. 

The reaction times in different conditions of the one-dimension natural background experiments are also shown in the left plot of Figure~\ref{figure_S12}.B. Although the differences are not statistically significant, the absolute increase in reaction time in $\Delta_{Sc}$ and $\Delta_{RD}$  is higher than in the other conditions, confirming that these variations need more processing time (note that the results are the average of five subjects only and increasing the sample size might lead to observe significant differences). In addition, $\Delta_{Po}=0$ and $\Delta_{RP}$ has the lowest reaction time increment (see the right plot of Figure~\ref{figure_S12}.B), meaning that these variations need less processing time in the human visual system.

\end{document}